%% file: main.tex
\documentclass[letterpaper, 10 pt, conference]{ieeeconf}

\IEEEoverridecommandlockouts                              % 
\usepackage{graphics} % for pdf, bitmapped graphics files
\usepackage{epsfig} % for postscript graphics files
\usepackage{amsmath} % assumes amsmath package installed
\usepackage{amssymb}  % assumes amsmath package installed
\usepackage{mathrsfs}
\usepackage{cite}
\usepackage{bm}
\usepackage{subcaption}
\usepackage{acronym}
\usepackage{paralist}
\usepackage{float}
\usepackage[dvipsnames]{xcolor}
\usepackage{epstopdf}
\usepackage{multicol}
\usepackage{tikz}
\usepackage{hyperref}
\usepackage{graphicx}
\usepackage{algorithm}
\usepackage{algpseudocode}
\usepackage{xcolor}
\usepackage{etoolbox} 

\usepackage{macros} % Adeel's custom defined macros

\usepackage{import}
\usepackage{xifthen}
\usepackage{pdfpages}
\usepackage{transparent}

\usepackage{./macros/rgsEnvironments}

\input{usercommands.tex} %Nan's custom defined here.
\newboolean{conf}
\setboolean{conf}{true}
\newcommand{\myifconf}[2]{\ifbool{conf}{#1}{#2}}
 \newcommand{\myblue}{\color{RoyalBlue}}
 \newcommand{\akh}{}

\makeatletter
\hypersetup{colorlinks=true}
\AtBeginDocument{\@ifpackageloaded{hyperref}
  {\def\@linkcolor{blue}
  \def\@anchorcolor{red}
  \def\@citecolor{red}
  \def\@filecolor{red}
  \def\@urlcolor{red}
  \def\@menucolor{red}
  \def\@pagecolor{red}
\begingroup
  \@makeother\`%
  \@makeother\=%
  \edef\x{%
    \edef\noexpand\x{%
      \endgroup
      \noexpand\toks@{%
        \catcode 96=\noexpand\the\catcode`\noexpand\`\relax
        \catcode 61=\noexpand\the\catcode`\noexpand\=\relax
      }%
    }%
    \noexpand\x
  }%
\x
\@makeother\`
\@makeother\=
}{}}
\makeatother

\makeatletter
\let\NAT@parse\undefined
\makeatother
\usepackage{url}
\usepackage[]{hyperref} %%
\urldef{\CombinedEmail}\url{{nanwang; ricardo}@ucsc.edu}
\urldef{\AdeelEmail}\url{adeel.akhtar@njit.edu}

\IEEEoverridecommandlockouts

\def\BibTeX{{\rm B\kern-.05em{\sc i\kern-.025em b}\kern-.08em
    T\kern-.1667em\lower.7ex\hbox{E}\kern-.125emX}}

\begin{document}

\title{\bf A Safe Hybrid Control Framework for Car-like Robot with Guaranteed Global Path-Invariance using a Control Barrier Function$^*$}

\author{Authors}

\author{Nan Wang$^{1}$ \qquad Adeel Akhtar$^{2}$ \qquad Ricardo G. Sanfelice$^{1}$
% <-this % stops a space
  \thanks{*Research by N. Wang and R. G. Sanfelice is partially supported by NSF Grants no. CNS-2039054 and CNS-2111688, by AFOSR Grants nos. FA9550-19-1-0169, FA9550-20-1-0238, FA9550-23-1-0145, and FA9550-23-1-0313, by AFRL Grant nos. FA8651-22-1-0017 and FA8651-23-1-0004, by ARO Grant no. W911NF-20-1-0253, and by DoD Grant no. W911NF-23-1-0158.}% <-this % stops a space
  % \thanks{Nan Wang, A. Akhtar, Eric Partika, and R. G. Sanfelice are with the Department of Electrical and Computer Engineering at the University of California at Santa Cruz, California, USA. \CombinedEmail}%%
  \thanks{$^{1}$N. Wang and R. G. Sanfelice are with the Department of Electrical and Computer Engineering at the University of California at Santa Cruz, California, USA. \CombinedEmail}
  \thanks{$^{2}$A. Akhtar is with the Department of Mechanical and Industrial Engineering at New Jersey Institute of Technology (NJIT), New Jersey, USA. \AdeelEmail}%%
}

\maketitle

%%%%%%%%%%%%%%%% Abstract %%%%%%%%%%%%%%%%%%%%%%%%%%%%%%%%%%
\begin{abstract}
This work proposes a hybrid framework for car-like robots with obstacle avoidance, global convergence, and safety, where safety is interpreted as path invariance, namely, once the robot converges to the path, it never leaves the path. Given \textit{a priori}  obstacle-free feasible path where obstacles can be around the path, the task is to avoid obstacles while reaching the path and then staying on the path without leaving it. The problem is solved in two stages. Firstly, we define a ``tight'' obstacle-free neighborhood along the path and design a local controller to ensure convergence to the path and path invariance. The control barrier function technology is involved in the control design to steer the system away from its singularity points, where the local path invariant controller is not defined. Secondly, we design a \mynne{hybrid control} framework that integrates this local path-invariant controller with any global tracking controller from the existing literature without path invariance guarantee, ensuring convergence from any position to the desired path, namely, global convergence. This framework guarantees path invariance and robustness to sensor noise. Detailed simulation results\myifconf{ affirm}{ and experimental validation on the OSOYOO Robot Car affirm} the effectiveness of the proposed scheme.

\end{abstract}
%%%%%%%%%%%%%%%%%%%%%%%%%%%%%%%%%%%%%%%%%%%%%%%%%%%%%%%%%%%%

%%%%%%%%%%%%%%%% Latex files for each funtions %%%%%%%%%%%%%
% \myac{This is Adeel's macro.}
% \mynn{This is Nan's macro.}

\input{tex/intro}

\input{tex/curves}
\input{tex/problem}
\input{tex/hybridmodel}
\input{tex/control}
\input{tex/results}
\input{tex/conclusion}
\input{tex/appendix}
%%%%%%%%%%%%%%%%%%%%%%%%%%%%%%%%%%%%%%%%%%%%%%%%%%%%%%%%%%%%

\vspace{-0.4cm}
%%%%%%%%%%%%%% Bibliographies %%%%%%%%%%%%%%%%%%%%%%%%%%%%%%
\bibliographystyle{IEEEtran}
\bibliography{myreferences}
%%%%%%%%%%%%%%%%%%%%%%%%%%%%%%%%%%%%%%%%%%%%%%%%%%%%%%%%%%%%

\end{document}

%% file: usercommands.tex
\newcommand{\reals}{\mathbb{R}}   

\newcommand{\pn}[1]{{#1}}
\newcommand{\nw}[1]{{#1}}
\newcommand{\ak}[1]{{#1}}%% color for changes made by Akhtar

\newcommand{\mynn}[1]{#1}
\newcommand{\mynnd}[1]{#1}
\newcommand{\mynne}[1]{#1}

\newcommand{\by}{\mynn{\delta_{y}}}

\newcommand{\agx}{\overline{x}}

%% file: tex/intro.tex
\section{Introduction}
\label{sec:introduction}

Desired motion of a nonholonomic system such as a car-like robot can be achieved by tracking a trajectory or following a path~\cite{NieFulMag10}. 
It is crucial to emphasize the fundamental distinction between a path and a trajectory~\cite{AkhNieWas2015}. A path represents a collection of points in the output space, independent of time, whereas a trajectory is a time-parameterized curve in the output space~\cite{AkhNie2011}. Importantly, a single path encompasses all associated trajectories. 
% We want to highlight that path and trajectory are fundamentally different~\cite{AkhNieWas2015}. A path is a set of points in the output space without a notion of time, while a trajectory is a curve in the output space parameterized by time~\cite{AkhNie2011}. Moreover, a path encapsulates the set of all corresponding trajectories. 
As highlighted in~\cite{Aguiar}, in trajectory tracking, \pn{there is a fundamental performance limitation that the smallest achievable tracking error is equal to the least amount of control energy needed to stabilize the error system}. \pn{While trajectory tracking methods may exhibit global convergence, this limitation prevents trajectory tracking methods from establishing the path invariance property, which guarantees that once the robot reaches the path, it remains on that path precisely and indefinitely \cite{LiNie2017}.} \pn{In environments with the presence of static obstacles, robots controlled by trajectory tracking controllers may deviate from their path, leading to collisions with obstacles. Consequently, such controllers are deemed \emph{unsafe} in this paper.} This can be overcome by switching the control objective to force the output to follow a path instead.
\ifbool{conf}{}{In this paper, we aim to formulate and solve a path-following problem for car-like robots.}

One of the advantages of considering a path-following problem is that this problem can be cast into a set stabilization problem, where the given path is treated as a set, and subsequently, the path can be made invariant~\cite{ConMagNie2010,AkhNieWas2015,NieMag2008}. 
% In simple words, if the path is made invariant, once the robot converges to the path, it stays on the path forever~\cite{LiNie2017,ConMagNie2010}. 
% A path-invariant controller performs better in terms of the deviation from the path when compared to a corresponding trajectory tracking controller~\cite{AkhNieWas2015}. 
However, one significant limitation of the existing path following controllers is that these controllers are local in the sense that the basin of attraction is not the entire \mynn{state} space~\cite{AkhNieWas2015}. 
% ~\cite{AkhNie2011,AkhNieWas2015,HlaNieWan11,NieMag04}. 
\begin{figure}[t]
    \centering
	\def\svgwidth{0.9\columnwidth}
	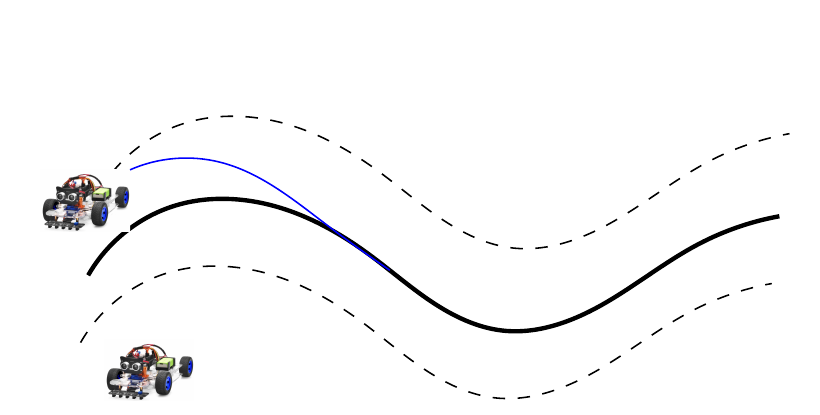

    \caption{\myifconf{Our proposed solution, shown in bold green and blue, guarantees global path invariance and robustness to sensor noise. In contrast, other solutions, shown in red and blue, fail to achieve both. LP-I stands for locally path invariant.}{Our proposed solution shown in bold green and blue guarantees globally path invariance and robustness to sensor noise. Other solutions fail to achieve both, shown in red and blue. LP-I is the abbreviation for locally path invariant.}}
    \vspace{-1.0cm}
    \label{fig:i}
\end{figure}
\myifconf{}{In other words, the existing path-following methods fail to guarantee global path invariance.} This limitation hampers the application of path-invariant control due to its lack of assured convergence, contingent upon the initial state of the car-like robots. Moreover, the path invariant controllers presented in~\cite{AkhNieWas2015,AkhNie2011} are not well defined for singularity points {\akh  where the decoupling matrix is not invertible} in the output space. Finally, since the system \myifconf{}{under study} has no potential energy, it violates Brockett's condition~\cite{BrockettsCondition-1983}, and any smooth time-invariant feedback cannot stabilize an equilibrium point{\akh~\cite{ReySchMcCKol1999,LewisMurray1999,ChenKla2017}}. 

To overcome these limitations, 
% \begin{figure*}[htbp]
%     \centering
%     % \includegraphics[width = \columnwidth]{Figures/NH.eps}
%     \incfig[2]{generaldiagram}
%     \caption{\ak{We need to either remove or significantly polish this figure.} The general framework of the proposed hybrid path-invariant path-tracking controller for self-driving robots.}
%     \label{fig:generalframework}
% \end{figure*}
we propose a hybrid \emph{globally path-invariant} control framework \pn{following the uniting control framework in \cite{San2021}}.
% Moreover, the resulting hybrid controller provides global converges and is robust to sensor noise~\cite{San2021}. 
Our hybrid control framework \nw{switches between} two controllers. The first controller is a local path-following controller that establishes path invariance if the robot is initialized in the neighborhood of the path. \nw{To establish the path invariance of this controller, we first extend the dynamic system by treating the control input as a state so as to add this state and its derivative into the state space. Then, by defining the neighborhood of the path in the extended state space and employing a traverse feedback linearization approach, we develop a locally path-invariant controller, albeit encountering singular points. To avoid reaching those singularity points, \ifbool{conf}{}{which may result in control failure,} we introduce a singularity filter for the local path-invariant controller, integrating a Control Barrier Function (CBF) \cite{ames2016control}.  The second controller can be any proper combination of motion planner and global trajectory tracking controller in the literature such as the sampling-based planning algorithms \cite{ma2015efficient} and pure pursuit tracking controller \cite{nascimento2018nonholonomic}, which is not necessarily path invariant, bringing the robot into the path's neighborhood in finite time.}
% Firstly, a trajectory, which is referred to as \emph{motion plan}  and connects the current position of the robot with the given path, is generated by the onboard planning module. 
% In addition to the classic graph search algorithm\cite{likhachev2005anytime} and those inspired by the artificial potential field \cite{huang2019motion}, 
% In recent years, the sampling-based algorithms \cite{ma2015efficient} have drawn much attention for their rapid exploration in solving high-dimensional planning problems. In this paper, the HyRRT algorithm in \cite{wang2022rapidly} is employed to generate an motion plan for the car-like robot. 
% Secondly, a global trajectory tracking controller is employed to track the motion plan, bringing the robot into the path's neighborhood in finite time. 
\ifbool{conf}{}{Once the robot enters the neighborhood of the given path, using a hybrid switching scheme, the control law {robustly} switches from the global trajectory tracking controller to the local path-invariant controller. While switching between two controllers may seem straightforward, the process \mynn{requires global convergence and robust switching}. }Following the uniting control framework in \cite{San2021}, the proposed hybrid control framework prevents the chattering between the two controllers to ensure global convergence.

To the authors' best knowledge, the proposed hybrid framework is the first control design for car-like robots that establishes global path invariance. In summary, we make the following contributions:
\ifbool{conf}{i) A path-following controller that guarantees local path invariance {and a barrier certificate to avoid singularities}~ (Lemma~\ref{lemma:invariance});
    % \item A pure pursuit controller that guarantees finite-time convergence to the neighborhood of the given path (Remark~\ref{r});
    ii) A hybrid framework that guarantees robust switching between the two controllers such that the resulting closed-loop system makes the given path \textit{globally invariant} (Theorem~\ref{theo:geometric-hybrid-cricle}); 
    % iii) Experiments demonstrate the performance of the proposed hybrid controller on a small-scale car \pn{when operating amidst real-world noise conditions}.
    iii) Despite the simplification of the robot dynamical system addressed in this work, the underlying theory is readily adaptable and extendable to other robotic systems.}{\begin{enumerate}
    \item A path-following controller that guarantees local path invariance {and a barrier certificate to avoid singularities}~ (Lemma~\ref{lemma:invariance});
    % \item A pure pursuit controller that guarantees finite-time convergence to the neighborhood of the given path (Remark~\ref{r});
    \item A hybrid framework that guarantees robust switching between the two controllers such that the resulting closed-loop system makes the given path \textit{globally invariant} (Theorem~\ref{theo:geometric-hybrid-cricle}); 
    \item Experiments demonstrate the performance of the proposed hybrid controller on a small-scale car \pn{when operating amidst real-world noise conditions}.
    \item \pn{Despite the simplification of the robot dynamical system addressed in this work, the underlying theory is readily adaptable and extendable to other robotic systems.}
    % , in a lab environment, 
\end{enumerate}}
% \ifbool{conf}{}{
% The remainder of the paper is organized as follows. 
Section~\ref{section:preliminaries} presents notation and preliminaries. Section~\ref{section:problem} presents the problem statement. Section~\ref{section:localcontrol} presents the local path-invariant controller for car-like robots. Section~\ref{sec:control} presents the hybrid control design. Sections~\ref{section:simulation} \myifconf{includes}{ and \ref{section:experiment}include}  the simulation \myifconf{results. More details are included in the report version \cite{wang2025hybrid}.}{ and experiment results.}
% }

% We will write introduction after sorting out the main sections.
% We need to structure the paper in somewhat the following manner:
% \begin{itemize}
%     \item Pretty much at the start of the paper we need to state informally and perhaps with the help of a picture our problem statement. 
%     \item Later we need to formally state our problem statement. 
%     \item Class of curves for which we're going to propose a global solution and the assumptions these curves satisfy.
%     \item We need to design two controller $\kappa_0$. and $\kappa_1$. 
%     \item We call the first controller $\kappa_0$ which will be a local path invariant controller that, pretty much, comes from my previous paper. So we need to state the main results (in the form of Lemma) from there. Also we need to explicitly characterize a region of attraction for the first controller. 
%     \item If we are outside that region, we use another controller, say $\kappa_1$, which, with the help of path planning brings the robot in the region of attraction. We need to formally write a lemma that guarantees that the system will avoid obstacles and get to the region of attraction in finite time.
%     \item A ``uniting" scheme that seamlessly joins these two controller. This will be stated in the from of a theorem (the main result of our paper) that guarantees global converge to the path and invariance. 
%     \item A short section on how we use path planning to achieve obstacle avoidance and pushing to system in the neighbourhood of the path.
    
% \end{itemize}

\section{Notation and Preliminaries}\label{section:preliminaries}
% \subsection{Notations}\label{sec: math prelim}
% \subsection{Notation and Definition}
% We use a standard notation throughout this paper. Specifically,
\subsection{Notation}
In this paper, $\Real$ denotes the set of real numbers, $\mathbb R_{>0}$ denotes the set of positive real numbers, and $\mathbb R_{\geq0}$ denotes the set of nonnegative real numbers. We use \ak{$\norm{\cdot}_p$} to denote the $p-$norm, and \ak{$\norm{\cdot}$} is used to denote the Euclidean norm. In addition, we write $\partial S$ for the boundary of the set $S$, $\textrm{int}(S)$ for its interior, \ak{and $\norm{x}_S = \inf_{y\in S}\norm{x-y}$ for the distance from the point $x\notin S$ to the set $S$}. The Lie derivative of a function $V:\mathbb R^n\rightarrow \mathbb R$ along a vector field $f:\mathbb R^n\rightarrow\mathbb R^n$ at a point $x\in \mathbb R^n$ is denoted as $L_fV(x) \triangleq \frac{\partial V}{\partial x} f(x)$. The notation $\dom f$ denotes the domain of \pn{the} function $f$. The components of a typical vector $x\in \Real^n$ is represented by $x = (x_1, x_2, \cdots,x_n)$. A parametric curve $\sigma$ is defined as $\sigma: \dom \sigma \to \Real^2$. When the curve is not closed, $\dom \sigma = \Real$. 
%{\myblue (AA: We want to represent the domain of a generic curve as $\mbb{D}$, but when the image of the curve is a close path, such as a circle, the domain would be periodic. Otherwise, we represent the domain with $\mathbb{D}$.)} 
% We denote by $\dot x(t) = \frac{\text{d}x(t)}{\partial t}$ the time derivative of function $x(t)$ and by $y'(s) = \frac{\text{d}y(s)}{\ds}$ the derivative of $y(s)$ with respect to the stretched time $s$.
For closed curves with finite length $L$, this means that
 $\dom \sigma = \mod{\Real}{L}$ and $\sigma$ is $L$-periodic, i.e., for any $\lambda \in \dom \sigma$, $\sigma(\lambda + L) = \sigma(L)$. By an open neighborhood of a set $\mathcal{K} \subset \mathbb{R}^{n}$, we mean an open set in $\mathbb{R}^{n}$ containing $\mathcal{K}$.
\subsection{Hybrid Systems}
\input{tex/preliminary_hybridmodel.tex}
\subsection{Finite-time Stability}
\input{tex/preliminary_stability.tex}

%% file: Figures/pdfs/hybridControlResults_a.pdf_tex
%% Creator: Inkscape 1.0.1 (c497b03c, 2020-09-10), www.inkscape.org
%% Accompanies image file 'hybridControlResults_a.pdf' (pdf, eps, ps)
\begingroup%
  \makeatletter%
  \providecommand\color[2][]{%
    \errmessage{(Inkscape) Color is used for the text in Inkscape, but the package 'color.sty' is not loaded}%
    \renewcommand\color[2][]{}%
  }%
  \providecommand\transparent[1]{%
    \errmessage{(Inkscape) Transparency is used (non-zero) for the text in Inkscape, but the package 'transparent.sty' is not loaded}%
    \renewcommand\transparent[1]{}%
  }%
  \providecommand\rotatebox[2]{#2}%
  \newcommand*\fsize{\dimexpr\f@size pt\relax}%
  \newcommand*\lineheight[1]{\fontsize{\fsize}{#1\fsize}\selectfont}%
  \ifx\svgwidth\undefined%
    \setlength{\unitlength}{393.44865261bp}%
    \ifx\svgscale\undefined%
      \relax%
    \else%
      \setlength{\unitlength}{\unitlength * \real{\svgscale}}%
    \fi%
  \else%
    \setlength{\unitlength}{\svgwidth}%
  \fi%
  \global\let\svgwidth\undefined%
  \global\let\svgscale\undefined%
  \makeatother%
  \begin{picture}(1,0.50543362)%
    \lineheight{1}%
    \setlength\tabcolsep{0pt}%
    \put(0,0){\includegraphics[width=\unitlength,page=1]{hybridControlResults_a.pdf}}%
    \put(0.71752378,0.00613565){\makebox(0,0)[lt]{\lineheight{1.25}\smash{\begin{tabular}[t]{l}desired path\end{tabular}}}}%
    \put(0,0){\includegraphics[width=\unitlength,page=2]{hybridControlResults_a.pdf}}%
    \put(0.77183726,0.06114139){\makebox(0,0)[lt]{\lineheight{1.25}\smash{\begin{tabular}[t]{l}neighborhood\end{tabular}}}}%
    \put(-0.00297847,0.38519233){\makebox(0,0)[lt]{\lineheight{1.25}\smash{\begin{tabular}[t]{l}LP-I controller\end{tabular}}}}%
    \put(0,0){\includegraphics[width=\unitlength,page=3]{hybridControlResults_a.pdf}}%
    \put(0.54407421,0.30671302){\makebox(0,0)[lt]{\lineheight{1.25}\smash{\begin{tabular}[t]{l}robust\\switching\end{tabular}}}}%
    \put(0.70178621,0.34266133){\makebox(0,0)[lt]{\lineheight{1.25}\smash{\begin{tabular}[t]{l}LP-I controller\end{tabular}}}}%
    \put(0,0){\includegraphics[width=\unitlength,page=4]{hybridControlResults_a.pdf}}%
    \put(0.56072258,0.47494526){\makebox(0,0)[lt]{\lineheight{1.25}\smash{\begin{tabular}[t]{l}global convergence\end{tabular}}}}%
    \put(0,0){\includegraphics[width=\unitlength,page=5]{hybridControlResults_a.pdf}}%
    \put(0.33140704,0.02173073){\makebox(0,0)[lt]{\lineheight{1.25}\smash{\begin{tabular}[t]{l}chattering\end{tabular}}}}%
    \put(0,0){\includegraphics[width=\unitlength,page=6]{hybridControlResults_a.pdf}}%
    \put(0.33125509,0.36680493){\makebox(0,0)[lt]{\lineheight{1.25}\smash{\begin{tabular}[t]{l}hysteresis\end{tabular}}}}%
    \put(0,0){\includegraphics[width=\unitlength,page=7]{hybridControlResults_a.pdf}}%
  \end{picture}%
\endgroup%

%% file: tex/preliminary_hybridmodel.tex
Following \cite{San2021}, a hybrid system $\mathcal{H}$ with inputs is modeled as 
\begin{equation}
\mathcal{H}: \left\{              
\begin{aligned}               
\dot{z} & = F(z, u)     &(z, u)\in C\\                
z^{+} & =  G(z, u)      &(z, u)\in D\\                
\end{aligned}   \right. 
\label{model:generalhybridsystem}
\end{equation}
where $z\in \reals^{n}$ is the state, $u\in \reals^{m}$ is the input, $C\subset \reals^{n}\times\reals^{m}$ is the flow set, $F: \reals^{n}\times\reals^{m} \to \reals^{n}$ is the flow map, $D\subset \reals^{n}\times\reals^{m}$ is the jump set, and $G:\reals^{n}\times\reals^{m} \to \reals^{n}$ is the jump map, respectively. The continuous evolution of $x$ is captured by the flow map $F$. The discrete evolution of $x$ is captured by the jump map $G$. The flow set $C$ collects the points where the state can evolve continuously. The jump set $D$ collects the points where jumps can occur.

% \begin{definition}[robust stability~\cite{San2021}]
% \label{def:robust-stability}
% Given a hybrid closed-loop system $\mc{H}$, a nonempty closed set $\mc{A}\subset \ms{M}$ and an open set $\mc{U}\subset \ms{M}$ such that $\mc{A} \subset \mc{U}$, the set $\mc{A}$ is said to be robustly stable for $\mc{H}$ on $\mc{U}$ if for every proper indicator function $\varpi$ of $\mc{A}$ on $\mc{U}$, every function $\beta \in \mc{KL}$ such that 
% \[
% \varpi(x(t,j)) \leq \beta(\varpi(x(0,0)), t+j)\quad \forall (t,j)\in\dom x 
% \]
% for the solutions to $\mc{H}$ from $\mc{U}$, and every continuous function 
% $\rho^{*}:\ms{M} \to \Real_{\geq 0}$ that is positive on $\mc{U} \setminus \mc{A}$, the following holds: for each compact set $K \subset \mc{U}$ and each $\epsilon >0$, there exists $\delta^{*} >0$ such that for each solution {$x_{\rho}$} the perturbed system $\mc{H}_{\rho}$ with $\rho = \delta^{*}\rho^{*}$, starting from $x_{\rho}(0,0) \in K$ satisfies 
% $$
% \varpi(x_{\rho}(t,j)) \leq \beta(\varpi(x_{\rho}(0,0)), t+j)+ \epsilon \quad   \forall (t,j)\in\dom x_{\rho}.
% $$
% \end{definition}

%% file: tex/preliminary_stability.tex
\ifbool{conf}{Consider the following continuous-time system
\begin{equation}\label{eq:differentialequation}
    \dot{x} = f(x)
\end{equation}
with state $x\in \mathbb{R}^{n}$. The solution to (\ref{eq:differentialequation}) starting from $x_{0}\in\mathbb{R}^{n}$ is defined as a locally absolutely continuous function $\phi: \dom\phi\to\mathbb{R}^{n}$ satisfying $\phi(0) = x_{0}$ and $\dot{\phi}(t) = f(\phi(t))$ for each  $t\in\dom\phi\subset\mathbb{R}_{\geq 0}$. We assume uniqueness of maximal solutions to (\ref{eq:differentialequation}).
Following~\cite{bhat2000finite}, finite-time stability can be defined as follows.
\begin{definition}[Finite-time stability]
Given (\ref{eq:differentialequation})
with state $x\in \mathbb{R}^{n}$, a nonempty closed set $\mathcal{K}\subset \mathbb{R}^{n}$ is finite-time stable for (\ref{eq:differentialequation}) if
\begin{enumerate}
    \item $\mathcal{K}$ is Lyapunov stable for (\ref{eq:differentialequation});
    \item there exists an open neighborhood $\mathcal{N}$ of $\mathcal{K}$ that is positively invariant  for (\ref{eq:differentialequation}) and a positive definite function $T: \mathcal{N}\to \mathbb{R}_{\geq 0}$, called the settling-time function, such that\pn{,} for each initial state $x_{0}\in \mathcal{N}$, each maximal solution $\phi$ to (\ref{eq:differentialequation}) from $x_{0}$ satisfies
    $|\phi(T(x_{0}))|_{\mathcal{K}} = 0$ and, \mynne{when} $x_{0}\in \mathcal{N}\backslash\mathcal{K}$,
    $|\phi(t)|_{\mathcal{K}} > 0 \quad \mynne{\forall t\in [0, T(x_{0}))}.$
\end{enumerate}
\end{definition}}{
Consider the following continuous-time system
\begin{equation}\label{eq:differentialequation}
    \dot{x} = f(x)
\end{equation}
with state $x\in \mathbb{R}^{n}$. The solution to (\ref{eq:differentialequation}) starting from $x_{0}\in\mathbb{R}^{n}$ is defined as a locally absolutely continuous function $\phi: \dom\phi\to\mathbb{R}^{n}$ satisfying $\phi(0) = x_{0}$ and $\dot{\phi}(t) = f(\phi(t))$ for each  $t\in\dom\phi\subset\mathbb{R}_{\geq 0}$. A solution $\phi$ to (\ref{eq:differentialequation}) is said to be \emph{maximal} if there does not exist another solution $\phi_{1}$ such that $\dom \phi$ is a proper subset of $\dom \phi_{1}$ and $\phi(t) = \phi_{1}(t)$ for all $t\in \dom \phi$. We assume  uniqueness of maximal solutions to (\ref{eq:differentialequation}).
\begin{assumption}\label{assu:unimax}
    For any $x_{0}\in\Real^{n}$, there exists an unique maximal solution to (\ref{eq:differentialequation}) starting from $x_0$.
\end{assumption}
Next, following \cite{bhat2005geometric}, we introduce positive invariance (also known as forward invariance), Lyapunov stability, and finite-time stability for (\ref{eq:differentialequation}).
%, for any initial state $x_{0}\in \mathbb{R}^{n}$, the system has a unique solution $\phi$ from $x_{0}$ defined on $[0, \infty)$.
\begin{definition}[Positive invariance]
Given (\ref{eq:differentialequation})
with state $x\in \mathbb{R}^{n}$ \mynne{satisfying Assumption~\ref{assu:unimax}}, a set $\mathcal{K}\subset\mathbb{R}^{n}$ is positively invariant for (\ref{eq:differentialequation}) if, for any initial state $x_{0}\in\mathcal{K}$, each solution to (\ref{eq:differentialequation}) starting from $x_{0}$, denoted $\phi$, satisfies $\phi(t)\in\mathcal{K}$ for each $t\in\dom \phi$.
\end{definition}
\begin{definition}[Lyapunov stability]
Given (\ref{eq:differentialequation}) with state $x\in \mathbb{R}^{n}$ \mynne{satisfying Assumption~\ref{assu:unimax}}, a nonempty closed set $\mathcal{A}\subset \mathbb{R}^{n}$ is Lyapunov stable for (\ref{eq:differentialequation}) if, for every open neighborhood $\mathcal{N}_{\epsilon}$ of $\mathcal{A}$, there exists an open neighborhood $\mathcal{N}_{\delta}$ of $\mathcal{A}$ such that, for any initial state $x_{0}\in\mathcal{N}_{\delta}$, each solution to (\ref{eq:differentialequation}) starting from $x_{0}$, denoted $\phi$, satisfies $\phi(t)\in\mathcal{N}_{\epsilon}$ for each $t\in\dom \phi$.
\end{definition}

Having defined positive invariance and Lyapunov stability, we proceed to define finite-time stability.
\begin{definition}[Finite-time stability]
Given (\ref{eq:differentialequation})
with state $x\in \mathbb{R}^{n}$ \mynnd{satisfying Assumption~\ref{assu:unimax}}, a nonempty closed set $\mathcal{K}\subset \mathbb{R}^{n}$ is finite-time stable for (\ref{eq:differentialequation}) if
\begin{enumerate}
    \item $\mathcal{K}$ is Lyapunov stable for (\ref{eq:differentialequation});
    \item there exists an open neighborhood $\mathcal{N}$ of $\mathcal{K}$ that is positively invariant  for (\ref{eq:differentialequation}) and a positive definite function $T: \mathcal{N}\to \mathbb{R}_{\geq 0}$, called the settling-time function, such that\pn{,} for each initial state $x_{0}\in \mathcal{N}$, each maximal solution $\phi$ to (\ref{eq:differentialequation}) from $x_{0}$ satisfies
    $|\phi(T(x_{0}))|_{\mathcal{K}} = 0$ and, \mynne{when} $x_{0}\in \mathcal{N}\backslash\mathcal{K}$,
    $|\phi(t)|_{\mathcal{K}} > 0 \quad \mynne{\forall t\in [0, T(x_{0}))}.$
\end{enumerate}
\end{definition}
}

%% file: tex/curves.tex
\subsection{Class of Curves}
\label{sec:curves}
Given a smooth curve $\mc{C}$ in $\Real^2$ without self intersections, the curve $\mc{C}$ has a regular parametric representation, namely,
\begin{equation}
\label{eq:general_path}
%\begin{aligned}
\sigma : \dom \sigma\rightarrow\mathbb{R}^{2}, \qquad
\lambda\mapsto{\left(\sigma_{1}(\lambda), \sigma_{2}(\lambda)\right)},
%\end{aligned}
\end{equation}
where $\sigma\subset \Real$ is at least twice continuously differentiable, i.e., $C^2$, and $\mc{C}= \image{(\sigma)}$.
Since $\sigma$ is regular, without losing generality, we assume it is
unit-speed parameterized, i.e., $\norm{\sigma^\prime} \equiv 1$, where $\sigma^\prime$ is the derivative of $\sigma$ with respect to the parameter $\lambda$. Consequently, the curve $\sigma$ is parameterized by its arc length; for details, see~\cite{Pres2010,AkhNieWas2015}.  For a unit-speed curve $\sigma$ with parameter $\lambda$, its curvature $K(\lambda)$ at the point $\sigma(\lambda)$ is defined to be  $\norm{\sigma^{\prime\prime}(\lambda)}$,  where $\sigma^{\prime\prime}$ is the second derivative of $\sigma$ with respect to the parameter $\lambda$.
\begin{assumption}[Implicit representation]
  The curve $\mathcal{C}\subset \Real^2$ has implicit representation $
  \gamma = \set{y \in W : s(y) = 0}$, 
  % {\myblue (AA: We used the small Greek letter $\gamma$ because later we ``lift'' this set from the output space to state space and then call it capital gamma $\Gamma$)} 
  where $s: \dom s \to \Real$ is a smooth function such that the {Jacobian of $s$ evaluated \mynn{at} each point on the path is not zero, i.e., $\D_y s \neq 0$
  for each $y \in \mathcal{C}$} and $\dom s \mynn{\subset}
  \Real^2$ is a set {consisting of an open neighborhood of the curve $\mathcal{C}$}.  
%   Moreover, there
%     exist two class-$\mathcal{K}_\infty$ functions $\alpha, \beta :
%     [0, \infty) \rightarrow [0, \infty)$ such that
% \begin{equation}
%   \left( \forall y \in W \right) \; \alpha{\left(
%       \|y\|_{\mathcal{C}} \right)} \leq \|s(y)\| \leq
%   \beta{\left(\|y\|_{\mathcal{C}} \right)}.
%  \label{eq:classK}
% \end{equation}
\label{ass:implicit}
\end{assumption}

Assumption~\ref{ass:implicit} assumes that the entire path is represented as
the zero-level set of the function $s$, at least locally. A simple example of such a curve is a unit circle with a parametric representation of $\lambda\mapsto(\cos \lambda,\sin\lambda)$ and an implicit representation of $s(y) = y_1^2 + y_2^2 - 1 = 0$,  {and $\dom s = \Real^2$. On the other hand, for $y = (y_1,y_2)\in\Real^2$, an $n$-th order polynomial in variable $y_1$ can be expressed as $y_2 = \sum_{i=0}^{n} a_iy_1^i$, where  scalars $a_i\in\Real$ and $y_1^i$ is the $i$-th power of $y_1$ for $i \in \{0, 1,..., n\}$. Moreover, the polynomial can be expressed implicitly as $s(y) = y_2 - \sum_{i=0}^{n} a_iy_1^i$, and in the form of a parametric curve as $\lambda\mapsto(\lambda,\sum_{i=0}^{n} a_i\lambda^i)$}. 

% {\myred We need to define curvature of the parametric curve $\sigma$.}

%% file: tex/problem.tex
%% Mathematical model
\section{Problem Statement}\label{section:problem}
\label{sec:mathematical_model}
Consider the kinematic car-like robot model in~\cite{AkhNieWas2015} 
\begin{equation}
\begin{aligned}
\label{eq:car_robot}
    \dot{x}_{1} &= v\cos x_{3}, \quad\quad \dot{x}_{2} = v\sin x_{3}, \\ \dot{x}_{3} &= \frac{1}{\ell}\tan x_{4}, \quad\quad \dot{x}_{4} = \omega,
\end{aligned}
\end{equation}
% \begin{aligned}
%   \dot{x}_{1} &= v\cos x_{3}, \dot{x}_{2} = v\sin x_{3}, \\ \dot{x}_{3} &= \frac{1}{\ell}\tan x_{4}, \dot{x}_{4} = \omega,
% \end{aligned}
% \begin{equation}
% \label{eq:car_robot}
%  \dot{x}=
%   \left[
%   \begin{array}{cc}
%       \cos{x_{3}} & 0\\
%       \sin{x_{3}} & 0\\
%       \frac{1}{\ell}\tan x_{4} & 0\\
%       0 & 1\\
%   \end{array} \right]\left[
%   \begin{array}{c}
%       v\\
%       \omega\\
%   \end{array}\right],
% \end{equation}
where $(x_1,x_2)\in\Real^2$ is the position in the two-dimensional plane, $x_3\in\Real$ is the orientation, the constant $\ell\in\mathbb R_{>0}$ is the length of the car-like robot, and $x_4\in\Real$ is the steering angle. For a constant upper bound $x^{{\max}}_4 \in (0,\pi/2)$, the steering angle has the following limits:
\begin{equation}
 |x_4| \leq x^{{\max}}_4.
\label{eq:constraint}
\end{equation}
The input $(v,\omega) \in \Real^{2}$ is the translational speed and angular velocity, respectively. The state $x = (x_{1}, x_{2}, x_{3}, x_{4})$ is assumed to be measurable, but we define the output of~\eqref{eq:car_robot} as the position of the car-like robot in the plane, given by
\begin{equation}
  \tilde{y} =\tilde{h}(x):={\left(x_1, x_2\right)}.
\label{eq:output}
\end{equation}
This output is used as feedback to the path-invariant controller.

\begin{assumption}
  Given the steering angle constraint~\eqref{eq:constraint}, the
  curvature of the curve $\sigma$ in~\eqref{eq:general_path}, denoted $K(\lambda)$, 
  satisfies
  $
  K^{\max}\eqdef \sup_{\lambda \in \nw{\dom \sigma}} K(\lambda) < \frac{1}{\ell}\tan{(x^{\max}_4)}
  $
% $
%  K(\lambda)
%  < \frac{1}{\ell}\tan{(x^{\text{max}}_4)},
%  \label{eq:curvature}
% $
for all $\lambda\in\dom \sigma$.
%Moreover, for a constant $\vrm > 0$, $x_{5}$ is such that $\vrm + \pn{x_5} \neq 0$ \pn{[NW: What is $x_{5}$?]}.
\label{ass:SteeringAngle}
\end{assumption}

Assumption~\ref{ass:SteeringAngle} ensures that the path is feasible,
in light of the steering angle constraint, for the car-like robot.

\begin{figure*}[htbp]
    \centering
	\def\svgwidth{1.73\columnwidth}
	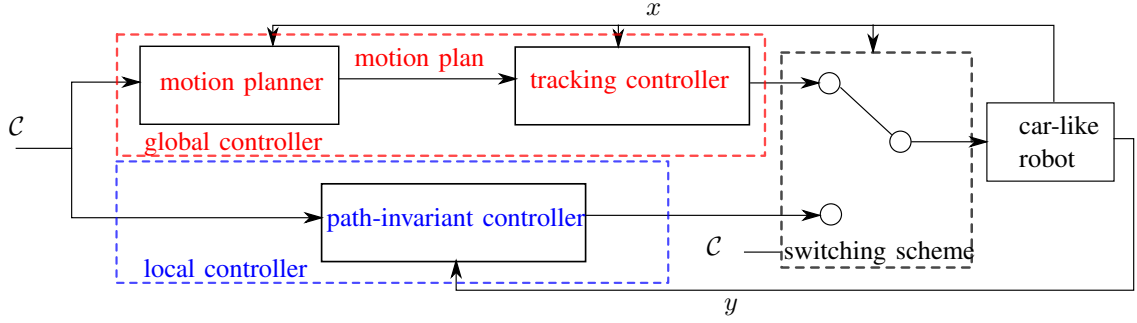

    \caption{The hybrid control diagram of the globally path-invariant control framework.}
    \label{fig:controldiagram}
    \vspace{-0.6cm}
\end{figure*}

\begin{problem}\label{problem:globalinvariance}
Given a car-like robot modeled as in~\eqref{eq:car_robot} with constraint in (\ref{eq:constraint}), and a curve $\mc{C}$ satisfying Assumptions~\ref{ass:implicit} and~\ref{ass:SteeringAngle}, design a controller $\kappa:\reals^{4}\to\reals^{2}$ such that, by applying $(v, w) = \kappa(x)$ to (\ref{eq:car_robot}) and with nonzero initial speed, the following holds: i) the output of (\ref{eq:car_robot}) converges to the curve $\mc{C}$ from any point in $\reals^{4}$, ii) the constraint in ~\eqref{eq:constraint} is satisfied,
% $$
% \mathcal{B} = \{x_{0}\in\reals^{4}: (\hat{x}(t|x_{0}), \kappa(\hat{x}(t|x_{0}))\notin \mathcal{S}\quad\forall t\in\mathbb{R}_{\geq 0}\}
% $$ 
% where $\hat{x}(t|x_{0})$ denotes the predicted state $x$ of the system (\ref{eq:car_robot}) controlled by $(v, w) = \kappa(x)$, denoted $P$, at time $t$ when starting from the initial state $x_{0}$, and $\mathcal{S}\subset\mathbb{R}^{6}$ is the set of singular states and inputs for $P$.
and iii) the system tracks the curve with a given non-zero speed, in the sense that the \pn{associated} set $\gamma$ is invariant. 
% Given a car-like robot modeled as in~\eqref{eq:car_robot}, and a curve $\mc{C}$ satisfying Assumptions~\ref{ass:implicit} and~\ref{ass:SteeringAngle}, design a controller such that output of the system converges to the curve $\mc{C}$ with a basin of attraction equal to $\Real^2$. Moreover, the system tracks the curve with a given non-zero speed, in the sense that the set $\gamma$ is invariant. 
\end{problem}
\ifbool{conf}{To \pn{solve} Problem \ref{problem:globalinvariance}, the hybrid control framework shown in Figure \ref{fig:controldiagram} \mynne{is employed}, comprising a global tracking controller that, powered by motion planning technology, drives the car-like robot into the neighborhood of the path, in conjunction with a local controller that guarantees path invariance of this neighborhood. A hysteresis-based switching scheme, following the uniting control framework introduced in \cite{San2021}, combines the two controllers.}{
 To \pn{solve} Problem \ref{problem:globalinvariance}, the hybrid control framework shown in Figure \ref{fig:controldiagram} \mynne{is employed}, comprising a global tracking controller that, powered by motion planning technology, drives the car-like robot into the neighborhood of the path, in conjunction with a local controller that guarantees path invariance of this neighborhood. Initially, the motion planner is tasked to generate a motion plan steering the robot to the neighbourhood of the desired path, where the locally path-invariant controller is effective. The global tracking controller is responsible for tracking the aforementioned motion plan, effectively guiding the robot into the designated neighborhood. Upon reaching this neighborhood, the hybrid control logic selects the locally path-invariant controller, which subsequently ensures path invariance. Conversely, if the robot exits some larger neighborhood of the path, designed to prevent chattering, the global controller takes over. This logic introduces \mynne{hysteresis} to prevent chattering when switching between the controllers, thereby ensuring robust switching.}
% The switching scheme, following the uniting control framework introduced in \cite{San2021}, is designed based on the \pn{distance between the position of the car-like robot and the path.}}

\section{Local Path-invariant Control Design}\label{section:localcontrol}
\label{sec:path-invariant-control-design}
In this section, we design a local controller to achieve the path invariance property. The \mynn{design} of \mynn{the} local controller is based on the transverse feedback linearization technique~\cite{NieMag04,AkhNieWas2015}. 
A necessary condition for feedback linearization is that the system possesses a well-defined vector relative degree~\cite{Isi95}. The next result shows that there does not exist any function in the output space of~\eqref{eq:car_robot} for which the system has a well-defined vector relative degree\myifconf{}{; \mynnd{see Definition \ref{def:vector_relative_deg}}}. 
\begin{lemma}
\label{lemm:no-vector-relative-degree}
    Given the system dynamics defined in~\eqref{eq:car_robot} and \mynnd{any} two smooth functions $(x_1,x_2)\mapsto A(x_1,x_2)\in\reals$ and $(x_1,x_2)\mapsto B(x_1,x_2)\in\reals$, the system~\eqref{eq:car_robot} does not have a well-defined vector relative degree. 
\end{lemma}
\myifconf{
\begin{proof}
    See \cite{wang2025hybrid}.
\end{proof}
}{\begin{proof}
    We can write the system~\eqref{eq:car_robot} as $\dot x = f(x) + g_1(x)v + g_2(x)\omega$, with $f(x) = [0,0,1/\ell\tan x_4,0]^\top$, $g_1(x) = [\cos x_3,\sin x_3, 0, 0]^\top$ and $g_2(x) = [0,0,0,1]^\top.$ {\akh For any smooth function $A$ and $B$, direct calculations result in $L_{g_{2}}A = 0$ and $L_{g_{2}}B = 0$, and the decoupling matrix given by
        $\begin{bmatrix}
            L_{g_{1}}A & L_{g_{1}}B\\
            L_{g_{2}}A & L_{g_{2}}B
        \end{bmatrix}$
        is always singular for any $A$ and $B$.} This implies that the system~\eqref{eq:car_robot} does not have a well-defined vector relative degree for any $A$ and $B$.
\end{proof}}

As a consequence of Lemma~\ref{lemm:no-vector-relative-degree}, it is not even possible to convert the given system~\mynne{\eqref{eq:car_robot}} into a \textit{partially linear} system~\mynne{~\cite[(4.26)]{Isi95}} using a \mynnd{pre-}feedback and coordinate transformation because the system does not possess a well-defined vector relative degree~\cite{Isi95}. 
To achieve a well-defined vector relative degree, similar to~\cite{AkhNieWas2015,AkhNie2011}, we perform dynamic extension by treating the control input $v\in\Real$ as a state and ``extend'' the system by adding two auxiliary states $x_5$ and $x_6$. 
% Let $x_5\in \mathbb{R}$ and $x_6\in \mathbb{R}$, be the two auxiliary states, and 
For a constant $\vrm \neq 0$, let  $v = \vrm + x_5$
% , where 
% $\zeta_1$ is
% the first state of our dynamic controller and 
 % is constant, $x_5 \in \reals\backslash \{-\vrm\}$, 
and $x_6 \in\Real$. \mynn{The dynamics of $x_{5}$ and $x_{6}$ are defined} as $\dot x_5 = x_6$ and $\dot{x}_6 = u_1$, where
$u_1$ is a new auxiliary input. In simple words, we have delayed the control input $v$ \mynn{via a double} integrator. Next, we relabel the input $\omega$ as $u_2$ and denote the state of the extended system as $\agx\eqdef (x_1, x_{2}, \cdots,x_6)$. \mynn{T}he extended system is defined as follows:
%
%
% In this section, we design a controller using transverse feedback linearization to achieve local path invariance ~\cite{NieMag04} for car-like robot. To achieve a well-defined vector relative degree for any curve in $\Real^2$, we perform dynamic extension~\cite{AkhNieWas2015} by adding two extra states of controller, \ak{$\zeta_1\in \mathbb{R}$ and $\zeta_2\in \mathbb{R}$, that capture input information}. Let $v = \vrm + \zeta_1$, where 
% % $\zeta_1$ is
% % the first state of our dynamic controller and 
% $\vrm \neq 0$ is
% constant. Furthermore, we define $\dot{\zeta_1} = \zeta_2$ and $\dot{\zeta}_2 = u_1$, where
% $u_1$ is a new auxiliary input. To simplify notation, henceforth, we do not distinguish between
% physical states of the system $(x_1,x_2, x_3, x_4)$ and states of the
% controller $(\zeta_1, \zeta_2)$. Let $x_5 \eqdef \zeta_1\in \reals\backslash \{-\vrm\}$, $x_6
%  \eqdef \zeta_2$ for some $\vrm > 0$. We denote the state of the extended system as $\agx\eqdef \col(x_1,\cdots,x_6)$. Therefore, the extended system is defined as follows:
%$ \dot{x}= f(x) + g_1(x)u_1 + g_2(x)u_2 $ where
\begin{equation}
\begin{aligned}
\label{eq:dynamic_car_robot}
 &\dot{\agx} = F_{P}(\agx, u)\eqdef f(\agx) + g_1(\agx)u_1 + g_2(\agx)u_2\\
  &  \mynn{:=}\left[
  \begin{array}{c}
      (\vrm + x_{5})\cos x_{3}\\
      (\vrm + x_{5})\sin x_{3}\\
      \frac{(\vrm + x_{5})}{\ell}\tan x_{4} \\
      0\\
      x_{6}\\
      0\\
  \end{array}\right]+ \left[
  \begin{array}{cc}
      0 \\
      0 \\
      0 \\
      0 \\
      0 \\
      1 \\
  \end{array}\right]u_1 +\left[
  \begin{array}{cc}
      0\\
      0\\
      0\\
      1\\
      0\\
      0\\
  \end{array}\right]u_2.
\end{aligned}
\end{equation}
The output of the extended model of the car-like robot is defined as
$
    {y} = h(\agx) \mynnd{:} = (x_1,x_2).
$
We lift the path $\gamma$ to the extended state space and construct the following set:
$$
\Gamma \eqdef \left(s \circ h\right)^{-1}(0) = \set{ \agx \in
  \mathbb{R}^6: s(h(\agx)) = 0},
\label{eq:lift}
$$
\mynn{with} the map $s$ \mynn{given as} in Assumption~\ref{ass:implicit}.
It should be noted that steering the output of the system to the curve is equivalent to steering the state of~\eqref{eq:dynamic_car_robot} to $\Gamma$. 
\mynn{For some positive real number $\by \in \left[0, \frac{\ell}{\tan{(x^{{\max}}_4)}}\right)$, where $\frac{\ell}{\tan{(x^{{\max}}_4)}}$ represents the minimal radius of the vehicle,} we construct a neighborhood of the path $\gamma$ \mynn{as follows}:
\begin{equation}
    \label{eq:nbh_set}
    {\mc{N}_{\gamma}^{\by} \eqdef \set{ y \in \Real^2 : \norm{y}_{\gamma} < \by}}.
\end{equation}
We lift this neighborhood \mynn{to the space of $\overline{x}$ and define}
\begin{equation}
    \label{eq:nbh_lift_set}
    {\mc{N}_{\Gamma}^{\by}} \eqdef \set{ \agx\in \Real^6 : \norm{\agx}_{\Gamma} < \by}.
\end{equation}

To solve Problem~\ref{problem:globalinvariance}, we need to satisfy two requirements: i) render the path invariant and \pn{(locally)} attractive, and ii) move along the curve \mynn{satisfying a given velocity profile}. 
%It should be noted that fulfilling the first requirement alone is trivial as one can turn off all the control inputs once the robot reaches the curve, and make the system stay on the path forever. 
However, fulfilling the path invariance requirement with motion along the curve with non-zero speed is challenging, as it is not easy to guarantee that once the robot reaches the path, it will never leave the path. In other words, we require a control law that renders a subset of $\mc{N}_\Gamma$
% \begin{equation}
%     \label{eq:nbh_lift_set}
%     {\mc{N}_{\Gamma}} \eqdef \set{ \agx\in \mc{N}^{\uparrow} : \vrm + x_5 \geq \delta}
% \end{equation}
invariant and \mynn{finite-time} attractive along the path, namely, satisfying $s(x) = 0$, where the condition $\vrm + x_5 = 0$ does not occur. To \mynne{satisfy requirements} i and ii, we exploit both the parametric and zero-level set representation of the path to construct two functions in the output space and invoke the local transverse feedback linearization design procedure~\cite{NieFulMag10,AkhNieWas2015}. This \mynn{requires} the formulation of a virtual output function.
%%%%%%%%%%%%%%%%%%%%%%%%%%%%%%%%%%%%%%%%%%%%%%%
% We treat the path following problem as a set stabilization problem and
% we follow the general approach of~\cite{NieFulMag10, HlaNieWan13}. In
% order to satisfy $\textbf{PF1}$ and $\textbf{PF2}$ we first stabilize
% the path following manifold $\Gamma^\star$. { Once the
%   path manifold has been stabilized we use the remaining freedom in the
%   control law to impose desired dynamics on the path and satisfy
%   $\textbf{PF3}$.}

\subsection{Virtual Output Function}
We construct a virtual output function using the definition of the path. \mynnd{Recalling the definition of} the \nbhd $\mc{N}_\gamma^{\by}$ in~\eqref{eq:nbh_set}, \mynn{there exists a small enough $\by$} such that if $y \in \mc{N}_{\gamma}^{\by}$ then there exists a unique $y^\star \in \gamma$ such
that $\norm{y}_\mc{\pn{\gamma}} = \norm{y - y^\star}$. This allows us to define the
function
\begin{equation}
\varpi : \; \mc{N}_{\gamma}^{\by} \rightarrow \nw{\dom \sigma},\; y \mapsto \varpi(y) :=\arg
\inf_{\lambda \in \nw{\dom \sigma}}\norm{y - \sigma(\lambda)}.
\label{eq:proj}
\end{equation}
This function is at least three times continuously differentiable. Next, we define \ak{the} virtual output function
\begin{equation}\label{eq:virtual}
\begin{aligned}
\hat{y} = \left[\begin{array}{c}\pi(\agx)\\\alpha(\agx) \end{array}\right] \eqdef
\left[\begin{array}{c}\varpi \circ h (\agx)\\s \circ h(\agx)\end{array}\right].
\end{aligned}
\end{equation}
% \begin{lemma}
% There does not exists any curve $\mc{C} \in \Real^2$ such that the corresponding path following output~\eqref{eq:virtual} has a well-defined vector relative degree for system~\eqref{eq:car_robot}. 
% \end{lemma}
% To overcome this problem let $v = \vrm + \zeta_1$, where $\zeta_1$ is
% the first state of our dynamic controller and $\vrm \neq 0$ is
% constant. We take the simplest possible structure for the control
% law~\eqref{eq:control} and let $\dot{\zeta_1} = \zeta_2$. In order to
% finish defining the control law we let $\dot{\zeta}_2 = u_1$ where
% $u_1$ is a new, auxiliary input. To simplify notation, henceforth we do not distinguish between
% physical states of the system $(x_1,x_2, x_3, x_4)$ and states of the
% controller $(\zeta_1, \zeta_2)$. Let $x_5 \eqdef \zeta_1$, $x_6
%  \eqdef \zeta_2$. Therefore the system we study has the form 
% %$ \dot{x}= f(x) + g_1(x)u_1 + g_2(x)u_2 $ where
% \begin{equation}
% \begin{aligned}
% \label{eq:dynamic_car_robot}
%  \dot{x}&= f(x) + g_1(x)u_1 + g_2(x)u_2\\&= \left[
%   \begin{array}{c}
%       (\vrm + x_{5})\cos x_{3}\\
%       (\vrm + x_{5})\sin x_{3}\\
%       \frac{(\vrm + x_{5})}{\ell}\tan x_{4}\\
%       0\\
%       x_{6}\\
%       0\\
%   \end{array}\right]+ \left[
%   \begin{array}{cc}
%       0 \\
%       0 \\
%       0 \\
%       0 \\
%       0 \\
%       1 \\
%   \end{array}\right]u_1 +\left[
%   \begin{array}{cc}
%       0\\
%       0\\
%       0\\
%       1\\
%       0\\
%       0\\
%   \end{array}\right]u_2
% \end{aligned}
% \end{equation}
The following results guarantee that the extended system in (\ref{eq:dynamic_car_robot}) with the output~\eqref{eq:virtual} has a well-defined vector relative degree (see \myifconf{\cite{Sastry}}{Definition~\ref{def:vector_relative_deg} in Appendix}) on $\mc{N}_{\Gamma}^{\by}$. It will be shown in the subsequent results that a well-defined vector-relative degree is key to guaranteeing the invertibility of a matrix that is required for the control law. \mynnd{Below, $\mc{N}^\star \eqdef \mc{N}_\Gamma^{\by} \setminus \set{\agx\in \Real^6 : \vrm + x_5  = 0}$.}
\begin{proposition}
\label{prop:relative_degree}
  Suppose Assumptions~\ref{ass:implicit} and~\ref{ass:SteeringAngle} hold. Then the extended model of the car-like robot~\eqref{eq:dynamic_car_robot} with output~\eqref{eq:virtual}
yields a well-defined vector relative degree of $\{3,3\}$ at each point {in} \akh{$\mc{N}^\star$}.
\end{proposition}
\myifconf{\begin{proof}
    See \cite{wang2025hybrid}.
\end{proof}}{\begin{proof}
The proof is similar to Lemma~{{III.1}} in~\cite{AkhNieWas2015}. 
  Let $x \in \mc{N}^\star$ be arbitrary. 
  % By definition of $\Gamma$ the output $h(x^\star)$ is on the path $\gamma$. Let $\lambda^\star \in \mathbb{D}$ be such that $h(x^\star) = \sigma(\lambda^\star)$.  
  By the definition of vector relative degree
  we must show that i) $ L_{g_1}L^i_f\pi(x) = L_{g_2}L^i_f\pi(x) =
  L_{g_1}L^i_f\alpha(x) = L_{g_2}L^i_f\alpha(x) = 0$ for $i \in \{0,
  1\}$ in a \nbhd of $x^\star$ and ii) the decoupling matrix
\begin{equation}
\label{eq:decoupling_matrix_general}
 D(x)=
  \left[\begin{array}{c c}
      L_{g_{1}}L_{f}^{2}\pi(x) & L_{g_{2}}L_{f}^{2}\pi(x)\\
      L_{g_{1}}L_{f}^{2}\alpha(x) & L_{g_{2}}L_{f}^{2}\alpha(x)\\
  \end{array}\right]
\end{equation}
is non-singular at $x\in\mc{N}^\star$. Since
\[
\DER{\pi(x)}{x_i} = \DER{\alpha(x)}{x_i} \equiv 0
\]
for {\akh each} $i \in \{3, 4, 5, 6\}$, direct calculations give
$L_{g_j}L^i_f\pi(x) = L_{g_j}L^i_f\alpha(x) = 0$ for $i \in \{0,1\}$,
  $j \in \{1, 2\}$. This satisfies the first condition of the vector relative degree.
{First, we write expressions for each entry of the decoupling matrix when $\alpha(x) = x_1^2 + x_2^2 - 1$ and $\pi(x) = \tan^{-1}(x_2/x_1)$. The closed-form expressions 
\begin{align*}
    L_{g_{1}}L_{f}^{2}\alpha(x) = 2 x_1 \cos x_3 + 2 x_2 \sin x_3,
\end{align*}
\begin{align*}
    L_{g_{2}}L_{f}^{2}\alpha(x) =\frac{2 (\vrm + x_5)^2 (\tan x_4^2 + 1) (x_2 \cos x_3 - x_1 \sin x_3)}{\ell},
\end{align*}
\begin{align*}
    L_{g_{1}}L_{f}^{2}\pi(x)=-\frac{x_2 \cos x_3 - x_1 \sin x_3}{x_1^2 + x_2^2},
\end{align*}
\begin{align*}
    L_{g_{2}}L_{f}^{2}\pi(x)=\frac{(\vrm + x_5)^2 (\tan x_4^2 + 1) (x_1 \cos x_3 + x_2 \sin x_3)}{\ell (x_1^2 + x_2^2)}.
\end{align*}
  For arbitrary functions, similar expressions can be computed from Maple. Next, we show that the decoupling
  matrix~\eqref{eq:decoupling_matrix_general} is non-singular at $x \in
  \mc{N}^\star$, we first find that
\begin{equation*}
\begin{aligned}
\displaystyle \det{(D(x))}=
\frac{(\vrm+x_{5})^2}{\ell\cos^{2}x_{4}}&\left(\partial_{x_{1}}\pi(x) \partial_{x_{2}}\alpha(x)\right.\\
 &\left.- \partial_{x_{2}}\pi(x) \partial_{x_{1}}\alpha(x) \right).
\end{aligned}
\end{equation*}
The only way for this determinant to vanish is if either (i) $\vrm =
-x_{5}$ or (ii)
$\left(\partial_{x_{1}}\pi(x) \partial_{x_{2}}\alpha(x) -
  \partial_{x_{2}}\pi(x)\partial_{x_{1}}\alpha(x)\right)$. Condition
(i) does not occur for $x \in \mc{N}^\star$. We now
argue that condition (ii) never occurs on the path because the term $\left(\partial_{x_{1}}\pi(x) \partial_{x_{2}}\alpha(x) -
  \partial_{x_{2}}\pi(x)\partial_{x_{1}}\alpha(x)\right)$ is the cross product of vectors
$\col(\partial_{x_{1}}\alpha,\partial_{x_{2}}\alpha)$ and
$\col(\partial_{x_{1}}\pi,\partial_{x_{2}}\pi)$. The cross product is zero if and only if these two vectors are linearly dependent. The construction of the maps $\alpha$ and $\pi$ in~\eqref{eq:virtual} along with Assumption~\ref{ass:implicit} guarantees that $\left(\partial_{x_{1}}\pi(x) \partial_{x_{2}}\alpha(x) -
  \partial_{x_{2}}\pi(x)\partial_{x_{1}}\alpha(x)\right) \neq 0$. This satisfies the second condition of the vector relative degree and the system has a well-defined vector relative degree of $\{3,3\}$ at each point \ak{in} {$\mc{N}^\star$}. 

% Returning to the expression for $\det{(D(x))}$, we have that
% \[
% \begin{aligned}
%   \sigma^\prime_{1}(\lambda^\star)\partial_{x_{2}}\alpha -
%   \sigma^\prime_{2}(\lambda^\star)\partial_{x_{1}}\alpha &=
%   \inner{R_{\frac{\pi}{2}}\D s^\top_{h(x^\star)}}{\sigma^\prime(\lambda^\star)}\\
%   &= k(\sigma(\lambda^\star))\inner{\sigma^\prime(\lambda^\star)}{\sigma^\prime(\lambda^\star)}\\
%   &= k(\sigma(\lambda^\star))\|\sigma^\prime(\lambda^\star)\|^2\\
%   &= k(\sigma(\lambda^\star)).
% \end{aligned}
% \]

}
\end{proof}}

The next result is a direct consequence of Proposition~\ref{prop:relative_degree}.
\begin{corollary}\label{cor:diffeo}
 The map $\mathscr{T}: \mc{N}^\star \rightarrow V \subset \Real^6 $ is defined as
  \begin{equation}
  \label{eq:diffeo}
        \mathscr{T}(\mynne{x}) := (L^{i-1}_f\pi(\mynne{x}),L^{i-1}_f\alpha(\mynne{x}))
  \end{equation}
%  
%   $T : \mc{N}_\Gamma{\myblue \setminus \set{\agx\in \Real^6 : \vrm + x_5 \neq 0} } \rightarrow
%   V \subset \Real^6$, where 
% \begin{equation}
% \label{eq:diffeo}
% T(x^\star) = (\eta_i,\xi_i) \eqdef (L^{i-1}_f\pi(x^\star),L^{i-1}_f\alpha(x^\star))
% \end{equation}
% \begin{equation}
% T(x^\star) = \left[\begin{array}{c}\eta_1\\\eta_2\\\eta_3\\ \xi_1\\\xi_2\\\xi_3\end{array}\right]
%  \eqdef
% \left[\begin{array}{c}\pi(x^\star)\\L_f\pi(x^\star)\\L^2_f\pi(x^\star)\\ \alpha(x^\star)\\L_f\alpha(x^\star)\\L^2_f\alpha(x^\star)\end{array}\right]
% \label{eq:diffeo}
% \end{equation} 
for $i\in\set{1,2,3}$ and, for any $\mynne{x} \in \mc{N}^\star $,
is a diffeomorphism onto its image $V\subset \Real^6$.
\end{corollary}
\myifconf{\begin{proof}
    See \cite{wang2025hybrid}.
\end{proof}}{\begin{proof}
% Let $\mc{N}^\star \eqdef \mc{N}_\Gamma^{\by} \setminus \set{\agx\in \Real^6 : \vrm + x_5  = 0}$. 
For each $x\in \mc{N}^\star$, the explicit expressions of $\pi$ and $\alpha$ and their Lie derivatives define transformation 
\begin{equation}
\mathscr{T}(x) = \left[\begin{array}{c}\eta_1\\\eta_2\\\eta_3\\ \xi_1\\\xi_2\\\xi_3\end{array}\right]
\eqdef 
\left[\begin{array}{c}\pi(x)\\L_f\pi(x)\\L^2_f\pi(x)\\ \alpha(x)\\L_f\alpha(x)\\L^2_f\alpha(x)\end{array}\right]
\label{eq:diffeo}
\end{equation}
in closed from. In order to show that~\eqref{eq:diffeo} is a diffeomorphism in a
  \nbhd of each point $x \in \mc{N}^\star  $ we appeal to the generalized inverse
  function theorem~\cite[pg. 56]{GuiPol74}. We must show that 1) for
  all $x \in \mc{N}^\star$, the Jacobian of $\mathscr{T}$ is an isomorphism, and 2)
  $\left.\mathscr{T}\right|_{\mc{N}^\star } : \mc{N}^\star  \to
  \mathscr{T}(\mc{N}^\star)$ is a diffeomorphism. An immediate consequence of
  Proposition~\ref{prop:relative_degree} and~\cite[Lemma 5.2.1]{Isi95} is
  that the first condition holds. To show that the second condition
  holds we explicitly construct the inverse of $\mathscr{T}$ restricted to
  $\mc{N}^\star$. On $\mc{N}^\star$, $\xi_1(x) = \xi_2(x) =
  \xi_3(x) = 0$ and simple calculations show that the inverse of $\mathscr{T}$
  restricted to $\mc{N}^\star$ is\footnote{The inverse is obtained
    under the assumption that the curve is arc-length parameterized.}
\[
\left[\begin{array}{c}x_1\\x_2\\x_3\\ x_4\\x_5\\x_6\end{array}\right]
= \left.\mathscr{T}\right|^{-1}_{\mc{N}^\star}(\eta, 0) =
\left[\begin{array}{c}\sigma_1(\eta_1)\\\sigma_2(\eta_1)\\\varphi(\eta_1)\\
    \arctan{\left(\ell K(\eta_1)\right)}\\\eta_2 - \vrm \\\eta_3\end{array}\right]
\]
where $\varphi : \dom\varphi \to \mod{\Real}{2\pi}$ is the map that
associates to each $\eta_1 \in \dom\varphi$ the angle of the tangent
vector $\sigma^\prime(\eta_1)$ to path $\gamma$ at $\sigma(\eta_1)$ and
$K$ is the signed curvature. The inverse
is clearly smooth which shows that $\left.\mathscr{T}\right|_{\mc{N}^\star}$
is a diffeomorphism onto its image.
\end{proof}}
Corollary~\ref{cor:diffeo} allows \mynne{to express} the system dynamics in $\xi$ and $\eta$ coordinates everywhere on $\mc{N}^\star$ and is given by
\begin{equation}
\label{eq:dynamic_quadrotor_new_coordinates}
\begin{aligned}
      \dot{\eta}_1 &= \eta_2\quad\dot{\eta}_2 = \eta_3\\
      \dot{\eta}_3 &= L_{f}^{3}\pi(x) +  \left(L_{g_1}L^2_{f}\pi(x)\right) u_1 +  \left(L_{g_2}L^2_{f}\pi(x)\right) u_2\\
      \dot{\xi}_1 &= \xi_2\quad\dot{\xi}_2 = \xi_3\\
      \dot{\xi}_3 &= L_{f}^{3}\alpha(x) +  \left(L_{g_1}L^2_{f}\alpha(x)\right) u_1 +  \left(L_{g_2}L^2_{f}\alpha(x)\right) u_2\\
\end{aligned}
\end{equation}

Next, we define a feedback $\kappa_{\mathrm{fb}}:\mc{N}^\star \to \Real^2 $, which is well-defined for all $x \in \mc{N}^\star$  by Proposition~\ref{prop:relative_degree}, as follows:
\begin{eqnarray}
\label{eq:regular_feedback_transformation}
 \left[\!\!\!
  \begin{array}{c}
      u_{1}\\
      u_{2}\\
  \end{array} \!\!\!\right]= \kappa_{\mathrm{fb}}(x):=D^{-1}(x)\left( \left[\!\!\!
  \begin{array}{c}
      -L_{f}^{3}\ak{\pi(x)}\\
      -L_{f}^{3}\ak{\alpha(x)}
  \end{array} \!\!\!\right]+
  \left[\!\!\!
  \begin{array}{c}
      v^{\parallel}\\
      v^{\pitchfork}\\
  \end{array} \!\!\!\right]
  \right),
\end{eqnarray}
where $(v^{\parallel},v^{\pitchfork}) \in \Real^{2}$ are
auxiliary control inputs and 
$
\label{eq:decoupling_matrix_general1}
 D(x)=
  \left[\begin{array}{c c}
      L_{g_{1}}L_{f}^{2}\pi(x) & L_{g_{2}}L_{f}^{2}\pi(x)\\
      L_{g_{1}}L_{f}^{2}\alpha(x) & L_{g_{2}}L_{f}^{2}\alpha(x)\\
  \end{array}\right].
$
On the set $\mc{N}^\star$, the coordinate transformation~\eqref{eq:diffeo} and the feedback~\eqref{eq:regular_feedback_transformation} converts the system into the linear system
\begin{subequations}
\label{eq:LTI_representation}
\begin{align}
    &\dot{\eta}_{1} =\eta_{2},\ \dot{\eta}_{2} =\eta_{3},\ \dot{\eta}_{3} =v^{\parallel}\label{eq:LTI_representationeta}\\
    &\dot{\xi}_{1} =\xi_{2},\  \dot{\xi}_{2} =\xi_{3}, \ \dot{\xi}_{3} =  v^{\pitchfork}\label{eq:LTI_representationxi}.
\end{align}
\end{subequations}
{
\begin{remark}
    It should be noted that the coordinate transformation ${\mathscr{T}}$ given in~\eqref{eq:diffeo} and the feedback law given in~\eqref{eq:regular_feedback_transformation} \nw{are} valid in the neighborhood of the path $\mc{N}_{\gamma}^{\by}$ if $\vrm + x_5 \mynnd{\neq} 0$. \akh{Moreover, $\eta_1$ is the path parameter, and on the path $\eta_2$ is the velocity of the robot.}
\end{remark}

% the condition $\vrm + x_5 \neq 0$ never occurs for the resulting closed-loop system.
}
\subsection{{Local Path-Invariant Controller with \nw{CBF-based} Singularity Filter}}
Next, we design a local path-invariant controller with a singularity filter that prevents $\vrm + x_5 = 0$, if { the system is initialized such that $\vrm + x_5 \geq \delta$ for some positive $\delta$.} We highlight that the linear system~\eqref{eq:LTI_representation} is geometrically equivalent to~\eqref{eq:dynamic_car_robot} at each point on the set $\mc{N}^\star$. Moreover, it consists of two chains of decoupled integrators and is valid everywhere in $\mc{N}^\star$. Therefore, we design a controller for~\eqref{eq:LTI_representation}.
% as it is linear
{We stabilize the origin of the $\xi$-subsystem by designing the \nw{local} controller $\kappa_\xi: \mc{N}_\Gamma^{\by} \to \Real$, which is given as
\begin{equation}
\label{eq:v_trans}
   v^\pitchfork = \kappa_{\xi}(\xi) =  -\sum_{i=1}^{3}k_{i} \sign(\xi_{i})|\xi_i|^{\beta_i},
\end{equation}
where for $i \in \{1, 2, 3\}$, $\beta_i>0$ is given as $\beta_1 = \frac{\beta}{2-\beta}$, $\beta_2 = \beta$ and $\beta_3 = 1$, for $\beta\in (1-\varepsilon, 1)$ where $0<\varepsilon<1$, and $k_i>0$ are such that the polynomial \myifconf{$
    \tilde s^3 + k_3\tilde s^2 + k_2\tilde s + k_1
$}{\begin{equation}\label{eq:hurwitz}
    \tilde s^3 + k_3\tilde s^2 + k_2\tilde s + k_1
\end{equation}} is Hurwitz (see \cite[Proposition 8.1]{bhat2005geometric}). 
% Before designing the controller for the $\eta$-subsystem, 
The transformed state $\eta_2$ is related to the speed of the robot, which is established in the following result.
\begin{proposition}
\label{prop:nonzero-speed}
{\akh For each $\bar{x}\in\mc{N}_\Gamma^{\by}$ satisfying $\inner{(\frac{\partial }{\partial x_1}\pi(\bar{x}),\frac{\partial }{\partial x_2}\pi(\bar{x}))}{\left(\cos \bar{x}^{0}_3,\sin \bar{x}^{0}_3\right)} >0$, where $\bar{x}^{0}_{i}$ denotes the $i$-th component of $\bar{x}^{0}$ for $i\in\{1, \ldots,6\}$, then $\eta_2 = 0$ if and only if $\mynne{\vrm + x_5} = 0$. Moreover, $\eta_2$ and $\mynne{\vrm + x_5}$ \mynne{have} the same sign.}
  % $\eta_2$ implies a nonzero robot's velocity, i.e., $\eta_2(t) \neq 0 \implies v + x_5 \neq 0$.} 
% For each trajectory solution $t\mapsto \agx := (x_{1}, x_{2}, x_{3}, x_{4}, x_{5}, x_{6})$ to (\ref{eq:dynamic_car_robot}) and its pointwise transformation $t\mapsto (\eta, \xi) := (\eta_{1}, \eta_{2}, \eta_{3}, \xi_{1}, \xi_{2}, \xi_{3})$ by $\mathscr{T}$ \mynnd{in (\ref{eq:diffeo})}, for each $t\in \dom(\eta, \xi) = \dom \agx$ such that  $\eta_2(t) = 0$, if and only if $v + x_5(t) \neq 0$.
  % $\eta_2$ implies a nonzero robot's velocity, i.e., $\eta_2(t) \neq 0 \implies v + x_5 \neq 0$.   
\end{proposition}
\myifconf{\begin{proof}
    See \cite{wang2025hybrid}.
\end{proof}}{\begin{proof}
    First, we argue that $v+x_5$ represents the magnitude of the velocity {\akh $(\dot x_1,\dot x_2)$} of the robot. From~\eqref{eq:dynamic_car_robot}, the $x_1$ and $x_2$ component of the velocity is given by $\dot x_1 = (v+x_5)\cos x_3$ and $\dot x_2 = (v+x_5)\sin x_3$. {\akh The} magnitude of the velocity is by $\sqrt{\dot x_1^2 + \dot x_2^2} = \sqrt{ (v+x_5)^2(\cos^2x_3 + \sin^2x_3) } = {\akh\norm{v+ x_5}}$. 
    
    {\akh Recall from~\eqref{eq:diffeo} that $\eta_1 = \pi(x)$. Moreover, it follows from~\eqref{eq:proj} and~\eqref{eq:virtual} that $\pi$ is only a function of $x_1$ and $x_2$. The derivative of $\eta_1$ is given by 
    \begin{align}
        \dot \eta_1 &= \eta_2 = \dot \pi(x)\\
             \eta_2 &= \partial_{x_1}\pi(x) \dot x_1 + \partial_{x_2}\pi(x) \dot x_2\\
                    & = \inner{(\partial_{x_1}\pi(x),\partial_{x_2}\pi(x))}{(\dot x_1,\dot x_2)}\\
                     & = (\vrm + x_5)\inner{(\partial_{x_1}\pi(x),\partial_{x_2}\pi(x))}{\left(\cos x_3,\sin x_3\right)}.
    \end{align}
    It follows from~\eqref{eq:proj} and~\eqref{eq:virtual} that $(\partial_{x_1}\pi(x),\partial_{x_2}\pi(x))$ is the velocity of the curve $\sigma$. Since $\sigma$ is regular, this implies that  $(\partial_{x_1}\pi(x),\partial_{x_2}\pi(x))$ is a non-zero vector. For any value of $x_3$, the vector $(\cos x_3,\sin x_3)$ is non-zero. Finally, by assumption $\inner{(\partial_{x_1}\pi(x),\partial_{x_2}\pi(x))}{\left(\cos x_3,\sin x_3\right)} >0$, which implies that $\eta_2 = 0$ if and only if $(\vrm + x_5) = 0$. Moreover, $\eta_2$ and $\mynne{\vrm + x_5}$ have the same sign. 
    }

    % Next, from~\eqref{eq:proj} it should be noted that the path parameter is $\lambda$, and by~\eqref{eq:virtual} and~\eqref{eq:diffeo}, the path parameter is equal to $\eta_1$. Moreover, $\eta_1$ represents the position of the robot on the path. Suppose the robot is moving along the path with some nonzero speed, i.e., $\dot\eta_1\neq 0$. By definition, $\eta_2 = \dot\eta_1$. The projection of $\eta_2$ on to the $x_1$ and $x_2$ axis is given by $\pr_{x_1}\eta_2$ and $\pr_{x_1}\eta_2$. Since, $\eta_2$ is nonzero, $\sqrt{ (\pr_{x_1}\eta_2)^2 + (\pr_{x_2}\eta_2)^2 } \neq 0$. Moreover, by the definition of $\eta_2$, it follows that $\pr_{x_1}\eta_2 = \dot x_1$ and $\pr_{x_2}\eta_2 = \dot x_2$. Hence, $\eta_2\neq 0 \implies \sqrt{ (\pr_{x_1}\eta_2)^2 + (\pr_{x_2}\eta_2)^2 } \neq 0 \implies v + x_5 \neq 0$.  
\end{proof}}
\begin{remark}
The condition on $\mc{N}_\Gamma^{\by}$, namely,
$\inner{(\frac{\partial }{\partial x_1}\pi(\bar{x}),\frac{\partial }{\partial x_2}\pi(\bar{x}))}{\left(\cos \bar{x}^{0}_3,\sin \bar{x}^{0}_3\right)} >0$, implies that if the robot is sufficiently close to the desired curve, then angle between the tangent vector of the curve representing the direction of curve and the velocity of the robot is less than $90^\circ$.
    % The transformation $\mathscr{T}$ in (\ref{eq:diffeo}) is not dependent on the input, as it only requires the state and the Lie derivative of the functions $\pi$ and $\alpha$ in (\ref{eq:virtual}) along the vector field $f$ in (\ref{eq:dynamic_car_robot}). Therefore, Proposition \ref{prop:nonzero-speed} holds regardless of input values.
\end{remark}
% \nw{Note that} the transformed state $\eta_1$ represents the position of the robot along the path, and $\eta_2$ represents the speed of the robot along the path and is equal to $\vrm + x_5$.

One can express the $\eta$ subsystem in the control affine form as $\dot \eta = \tilde{f}(\eta) + \tilde{g}(\eta)v^\parallel$, where $\tilde{f}\nw{(\eta)} = (\eta_2,\eta_3,0)$ and $\tilde{g}\nw{(\eta)} = (0,0,1)$. \nw{To} follow the path with a given reference velocity ${\eta}^{\mathrm{ref}}_{2}\in\reals$ and its derivative $\eta^{\mathrm{ref}}_{3}\in\reals$, we construct the Lyapunov function $V: \reals\times\reals\times \reals\times\reals\to \reals_{\geq 0}$ as follows:
$
V(\eta_{2}, \eta_{3}, {\eta}^{\mathrm{ref}}_{2}, \eta^{\mathrm{ref}}_{3}) = 1/2(\eta_2 - {\eta}^{\mathrm{ref}}_{2})^2 + 1/2(\eta_3 - { \eta}^{\mathrm{ref}}_{3})^2
$ and apply any control input from the set-valued map \nw{$K_{\mathrm{clf}}: \reals\times \reals\times\reals\times\reals\rightrightarrows \reals$} at the current \mynnd{$\eta_{2}, \eta_{3}$} and reference ${\eta}^{\mathrm{ref}}_{2}$ and $\eta^{\mathrm{ref}}_{3}$:
\myifconf{$
    K_{\mathrm{clf}}(\mynnd{\eta_{2}, \eta_{3}}, {\eta}^{\mathrm{ref}}_{2}, \eta^{\mathrm{ref}}_{3}) \eqdef\{v^{\parallel}\in\Real: L_{\tilde f}V(\mynnd{\eta_{2}, \eta_{3}}, {\eta}^{\mathrm{ref}}_{2}, \eta^{\mathrm{ref}}_{3})
    + L_{\tilde g}V(\mynnd{\eta_{2}, \eta_{3}}, {\eta}^{\mathrm{ref}}_{2}, \eta^{\mathrm{ref}}_{3})v^{\parallel}
    \leq {\akh -}\beta(V(\mynnd{\eta_{2}, \eta_{3}}, {\eta}^{\mathrm{ref}}_{2}, \eta^{\mathrm{ref}}_{3})) \}
$}{\begin{equation}
\begin{aligned}
    \label{eq:CLF}
    &K_{\mathrm{clf}}(\mynnd{\eta_{2}, \eta_{3}}, {\eta}^{\mathrm{ref}}_{2}, \eta^{\mathrm{ref}}_{3}) \eqdef\{v^{\parallel}\in\Real: L_{\tilde f}V(\mynnd{\eta_{2}, \eta_{3}}, {\eta}^{\mathrm{ref}}_{2}, \eta^{\mathrm{ref}}_{3})\\
    &+ L_{\tilde g}V(\mynnd{\eta_{2}, \eta_{3}}, {\eta}^{\mathrm{ref}}_{2}, \eta^{\mathrm{ref}}_{3})v^{\parallel}
    \leq {\akh -}\beta(V(\mynnd{\eta_{2}, \eta_{3}}, {\eta}^{\mathrm{ref}}_{2}, \eta^{\mathrm{ref}}_{3})) \}
\end{aligned}
\end{equation}}
to track the desired speed profile ${\eta}^{\mathrm{ref}}_{2}$ and $\eta^{\mathrm{ref}}_{3}$, where $\beta$ is a class $\mc{K}$ function. Next, we \nw{construct} a barrier function 
$
b: \Real^3 \to \Real
$ as 
$
b(\eta) \eqdef \delta - \eta_2
$ to guarantee that $\vrm + x_5 > \delta$, for some positive $\delta$.
% In other words, on the set $\mc{N}_\Gamma$, $\eta_2 > \delta$. 
\nw{The set \begin{equation}
\label{eq:set-S1}
 S_1 \eqdef \set{ \eta \in \Real^3 : b(\eta) \leq 0}   
\end{equation} cannot be forward invariant by selecting proper input $v^\parallel$ because the system has a relative degree two for the barrier function $b$ and the rate of change of $b$ along the vector fields $\tilde g$ is zero. Similar to~\cite{XiaBelCas2022}, let $\psi_0(\eta) \eqdef b(\eta)$ and $\psi_1(\eta) \eqdef \dot\psi_0{\eta} + \beta(\psi_0(\eta))$, where $\beta$ is a class $\mc{K}$ function. Next, using $\psi_1(\eta)$
we construct a second set 
\begin{equation}
\label{eq:set-S2}
S_2 \eqdef \set{ \eta \in \Real^3 : \psi_1(\eta) \leq 0}.   
% S_2 \eqdef \set{\eta \in \Real^3 : b(\eta) + L_{\tilde f}b(\eta) \leq 0 }.
\end{equation}
It should be noted that by Corollary~\ref{cor:diffeo}, the transformation $\mathscr{T}$ is a local diffeomorphism. Therefore the sets $S_1$ and $S_2$ can be expressed in $x$-coordinates by applying the inverse transformation, i.e., $\mathscr{T}^{-1}$.
By selecting a control input from the set-valued map $K_{\mathrm{cbf}}: \reals^{3}\to \reals$  at the current $\eta$ as follows:
\myifconf{$
        K_{\mathrm{cbf}}(\eta) \eqdef \{ v^{\parallel}\in\Real : L_{\tilde f}^2b(\eta) + L_{\tilde f}b(\eta) +L_{\tilde g}L_{\tilde f}b(\eta)v^{\parallel}
    \leq -\beta_k( \psi_1(\eta) )  \},
$}{\begin{equation}\label{eq:CBF}
\begin{aligned}
        K_{\mathrm{cbf}}(\eta) &\eqdef \{ v^{\parallel}\in\Real : L_{\tilde f}^2b(\eta) + L_{\tilde f}b(\eta) +L_{\tilde g}L_{\tilde f}b(\eta)v^{\parallel}  \\ 
    &\leq -\beta_k( \psi_1(\eta) )  \},
\end{aligned}
\end{equation}}
it will be shown in the following result that the set $S_1 \cap S_2$ is forward invariant.} Finally, to track the desired speed profile $({\eta}^{\mathrm{ref}}_{2}, \eta^{\mathrm{ref}}_{3})$ and guarantee that the system trajectories never reach the singularity point, i.e., $\vrm + x_5  = 0$, the control input is selected 
% from the set $K_{\mathrm{clf}}  \cap K_{\mathrm{cbf}}$. In other words, the controller $\kappa_\eta: \mc{N}_{\Gamma} \subset \Real^3 \to \Real$ such that 
\nw{as $\kappa_\eta(\eta, {\eta}^{\mathrm{ref}}_{2}, \eta^{\mathrm{ref}}_{3}) \in K_{\mathrm{clf}}(\eta, {\eta}^{\mathrm{ref}}_{2}, \eta^{\mathrm{ref}}_{3})\cap K_{\mathrm{cbf}}\nw{(\eta)}$}. In summary, for a given reference velocity profile ${\eta}^{\mathrm{ref}}_{2}$ and $ \eta^{\mathrm{ref}}_{3}$ that the robot is required to follow, we design the following controller:
$
    \kappa_0 : \mc{N}^{\by}_\Gamma\times\reals\times\reals \to \Real^{2}
$
such that, for {\akh each} $(\xi,\eta)\in \mc{N}_\Gamma^{\by}$,
\begin{equation}
\label{eq:kappa_0}
\begin{aligned}
     (\xi,\eta, {\eta}^{\mathrm{ref}}_{2}, \eta^{\mathrm{ref}}_{3}) &\mapsto \kappa_0(\xi,\eta, {\eta}^{\mathrm{ref}}_{2}, \eta^{\mathrm{ref}}_{3})\\
     &\eqdef  (\kappa_\xi(\xi), \kappa_\eta(\eta, {\eta}^{\mathrm{ref}}_{2}, \eta^{\mathrm{ref}}_{3})).
     % \left[ \begin{array}{c}
         % v^\pitchfork \\
         % v^\parallel 
    % \end{array}\right] \\
    % &= \left[ \begin{array}{c}
          % -\sum_{i=1}^{3}k_{i} \sign(\xi_{i})|\xi_i|^{\beta_i} \\
          % k_{4}(\eta_{2}-{\eta}^{\mathrm{ref}})+
      % k_{5}(\eta_{3}- \dot{\eta}^{\mathrm{ref}}) 
    % \end{array}\right],    
\end{aligned}
\end{equation}
% where for $i \in \{1, 2, 3\}$, $\beta_i>0$ is given as $\beta_1 = \frac{\beta}{2-\beta}$, $\beta_2 = \beta$ and $\beta_3 = 1$, for $\beta\in (1-\varepsilon, 1)$ where $0<\varepsilon<1$, and $k_i>0$ are such that $\tilde s^3 + k_3\tilde s^2 + k_2\tilde s + k_1$ is Hurwitz (see \cite[Proposition 8.1]{bhat2005geometric}). Moreover, we select $k_{4} > 0$ and $k_5 >0$ using pole placement, and the {\myblue parameter ${\dot\eta}^{\mathrm{ref}}$ is the derivative of the reference velocity along the curve}. 
% {\myblue
% \begin{definition}[Feasibility Condition]
%    A control policy $\kappa_\eta : \mc{N}_\Gamma^{\by}\times\reals\times\reals \to \Real$ for the closed-loop system $\dot \eta = \tilde{f}(\eta) + \tilde{g}(\eta)\kappa_\eta(\eta, {\eta}^{\mathrm{ref}}_{2}, \eta^{\mathrm{ref}}_{3})$ is feasible if it satisfies both the higher-order control barrier function constraint~\eqref{eq:CBF} and the control Lyapunov function constraint~\eqref{eq:CLF} \nw{for each point in $\mc{N}_\Gamma^{\by}\times\reals\times\reals$}. In other words, \nw{for each $\eta\in\mc{N}_\Gamma^{\by}$, ${\eta}^{\mathrm{ref}}_{2}\in\reals$ and  ${\eta}^{\mathrm{ref}}_{3}\in\reals$,} 
%  $K_{\mathrm{clf}}\nw{(\eta, {\eta}^{\mathrm{ref}}_{2}, {\eta}^{\mathrm{ref}}_{3})} \cap K_{\mathrm{cbf}}\nw{(\eta)} \neq \emptyset$.
% \end{definition}
% }

Then, we are ready to present the local path invariance result.
% We want to highlight that $\eta_1$ and $\eta_2$ are the position and velocity of the robot along the given path, respectively, as defined in~\eqref{eq:diffeo}. In this setting, the velocity of the robot is the control variable.
% 
% \begin{assumption}
% \label{ass:feasible_control}
%     \nw{For any $(\xi,\eta)\in \mc{N}_\Gamma$}, ${\eta}^{\mathrm{ref}}_{2}\in\reals$ and $\eta^{\mathrm{ref}}_{3}\in\reals$, the set of feasible control inputs in nonempty, i.e., $K_{\mathrm{clf}}\nw{(\eta, {\eta}^{\mathrm{ref}}_{2}, {\eta}^{\mathrm{ref}}_{3})} \cap K_{\mathrm{cbf}}\nw{(\eta)} \neq \emptyset$.
% \end{assumption}
  \begin{lemma}
  \label{lemma:invariance}
  \mynne{For each initial state $\bar{x}^{0} \in \mc{N}_\Gamma^{\by}$ satisfying i) the heading condition: $\inner{(\frac{\partial }{\partial x_1}\pi(\bar{x}),\frac{\partial }{\partial x_2}\pi(\bar{x}))|_{\bar{x}=\bar{x}^{0}}}{\left(\cos \bar{x}^{0}_3,\sin \bar{x}^{0}_3\right)} >0$; ii) and the velocity condition: $\vrm + \bar{x}^{0}_5 > \delta$ for some arbitrarily small positive $\delta\in\reals_{>0}$, where $\bar{x}^{0}_{i}$ denotes the $i$-th component of $\bar{x}^{0}$ for $i\in\{1, \ldots,6\}$,}
  {\akh then
  % for each initial conditions $\agx^{0}\in \mc{N}_\Gamma^{\by}$$
  % \setminus \{\agx\in\Real^6 : \vrm + x_5 > \delta\}
  the closed-loop system obtained by applying the controllers $\kappa_0$ given in~\eqref{eq:kappa_0}, to~\eqref{eq:LTI_representation}, renders the set
  $
  \Gamma^\star \eqdef \set{\agx\in\Real^6 : \alpha(\agx) = \dot\alpha (\agx)= \ddot\alpha(\agx) = 0}
  $
  finite-time stable with basin of attraction $\mc{N}^{\star}$  in~\eqref{eq:nbh_lift_set} and forward invariant. Moreover, the trajectories of the closed-loop system remain safe in the sense that the set $S_1 \cap S_2$ defined by~\eqref{eq:set-S1} and~\eqref{eq:set-S2} is forward invariant. Furthermore, \nw{for each $\eta\in\mc{N}_\Gamma^{\by}$, ${\eta}^{\mathrm{ref}}_{2}\in\reals$ and  ${\eta}^{\mathrm{ref}}_{3}\in\reals$,} 
 $K_{\mathrm{clf}}\nw{(\eta, {\eta}^{\mathrm{ref}}_{2}, {\eta}^{\mathrm{ref}}_{3})} \cap K_{\mathrm{cbf}}\nw{(\eta)} \neq \emptyset$.} 
  \end{lemma}
\myifconf{\begin{proof}
    See \cite{wang2025hybrid}.
\end{proof}}{\begin{proof}
  We first utilize Proposition 8.1 from \cite{bhat2005geometric} (refer to Proposition \ref{prop:linearfinitetime} in the Appendix) to demonstrate that zero is a globally finite-time-stable equilibrium for the subsystem (\ref{eq:LTI_representationxi}) under the feedback control in (\ref{eq:v_trans}), namely, the feedback controller in~\eqref{eq:kappa_0} steers all the $\xi$-states to converge to zero in finite time. Note that  for $i \in \{1, 2, 3\}$, $k_{i}$ is designed such that (\ref{eq:hurwitz}) satisfies the Hurwitz condition, and the feedback control in (\ref{eq:v_trans}) conforms to the structure of (\ref{eq:xicontrol}) with $n = 3$. Therefore, all the conditions in \cite[Proposition 8.1]{bhat2005geometric} are met, thereby establishing the finite-time stability of the origin for (\ref{eq:LTI_representationxi}).

  % Since~\eqref{eq:LTI_representation} is a linear system, by~\cite[Proposition 8.1]{bhat2005geometric} the controller~\eqref{eq:kappa_0} forces all the $\xi$-states to converge to zero in finite time. 
  From the definition of the coordinate transformation $\mathscr{T}$ given in~\eqref{eq:diffeo} $\xi_1$ is the zero-th Lie derivative of $\alpha(\agx)$, i.e., $\xi_1 = L_f^0\alpha(\agx) = \alpha(\agx)$. The first and the second Lie derivatives of $\alpha(\agx)$ defines $\xi_2$ and $\xi_3$ states, respectively, i.e., $\xi_2 = L_f\alpha(\agx) = \dot\alpha(\agx)$ and $\xi_3 = L_f^2\alpha(\agx) = \ddot\alpha(\agx)$.
  %
  % Moreover, as defined in~\eqref{eq:diffeo}, $\xi_1 = \alpha(\agx)$, $\xi_2 = \dot \alpha(\agx)$, and $\xi_3 = \ddot\alpha(\agx)$.
  Therefore by~\cite[Proposition 8.1]{bhat2005geometric}, the set $\Gamma^\star = \set{(\xi,\eta)\in\Real^6: \alpha(\agx) = \dot\alpha(\agx) = \ddot\alpha(\agx)}$ is finite-time stable and invariant for the closed-loop system~\eqref{eq:LTI_representation}. It follows from Corollary~\ref{cor:diffeo} that the basin of attraction of $\kappa_0$ is $\mc{N}_\Gamma^{\by}$. 
  To prove that the trajectories of the closed-loop system never get arbitrarily close to the point $\vrm + x_5 = 0$ if the initial condition belongs to the set $\mc{N}_\Gamma^{\by} \setminus \{\agx\in\Real^6 : \vrm + x_5 > \delta\}$. We note that by Proposition~\ref{prop:nonzero-speed}, it is sufficient to prove that $\eta_2 > 0$. One can readily verify that by~\cite[Definition 4]{XiaBelCas2022}, the candidate barrier function $b = \delta - \eta_2$ is a high-order control barrier function. Moreover, by~\cite[Theorem 2]{XiaBelCas2022} there exists a feasible control input in the set $K_{\mathrm{clf}}\nw{(\eta, {\eta}^{\mathrm{ref}}_{2}, {\eta}^{\mathrm{ref}}_{3})} \cap K_{\mathrm{cbf}}\nw{(\eta)} $ such that the set $S_1 \cap S_2$ defined by~\eqref{eq:set-S1} and~\eqref{eq:set-S2} is forward invariant.
  \end{proof}}
  \begin{remark}
     The invariance guarantee provided by Lemma~\ref{lemma:invariance} is twofold. First, it guarantees that the given path is locally attractive and forward-invariant, which means that the system converges to the path and then never leaves the path. Since by assumption, the path does not have obstacles, and by selecting a ``tight" obstacle-free neighborhood, one can guarantee safety. Second, it certifies in the neighborhood of the path the singularity condition ($\vrm + x_5 = 0$) will never occur by establishing the forward invariance of the set $S_1 \cap S_2$. 
  \end{remark}

}

%% file: Figures/pdfs/framework_globalpathinvariant.pdf_tex
%% Creator: Inkscape 1.3.2 (091e20e, 2023-11-25), www.inkscape.org
%% Accompanies image file 'framework_globalpathinvariant.pdf' (pdf, eps, ps)
\begingroup%
  \makeatletter%
  \providecommand\color[2][]{%
    \errmessage{(Inkscape) Color is used for the text in Inkscape, but the package 'color.sty' is not loaded}%
    \renewcommand\color[2][]{}%
  }%
  \providecommand\transparent[1]{%
    \errmessage{(Inkscape) Transparency is used (non-zero) for the text in Inkscape, but the package 'transparent.sty' is not loaded}%
    \renewcommand\transparent[1]{}%
  }%
  \providecommand\rotatebox[2]{#2}%
  \newcommand*\fsize{\dimexpr\f@size pt\relax}%
  \newcommand*\lineheight[1]{\fontsize{\fsize}{#1\fsize}\selectfont}%
  \ifx\svgwidth\undefined%
    \setlength{\unitlength}{726.44516832bp}%
    \ifx\svgscale\undefined%
      \relax%
    \else%
      \setlength{\unitlength}{\unitlength * \real{\svgscale}}%
    \fi%
  \else%
    \setlength{\unitlength}{\svgwidth}%
  \fi%
  \global\let\svgwidth\undefined%
  \global\let\svgscale\undefined%
  \makeatother%
  \begin{picture}(1,0.27912599)%
    \lineheight{1}%
    \setlength\tabcolsep{0pt}%
    \put(0,0){\includegraphics[width=\unitlength,page=1]{framework_globalpathinvariant.pdf}}%
    \put(0.4625523,0.20189188){\makebox(0,0)[lt]{\lineheight{1.25}\smash{\begin{tabular}[t]{l}\textcolor{red}{tracking controller}\end{tabular}}}}%
    \put(0.28150833,0.07813392){\makebox(0,0)[lt]{\lineheight{1.25}\smash{\begin{tabular}[t]{l}\textcolor{blue}{path-invariant controller}\end{tabular}}}}%
    \put(0,0){\includegraphics[width=\unitlength,page=2]{framework_globalpathinvariant.pdf}}%
    \put(0.89771839,0.16065073){\makebox(0,0)[lt]{\lineheight{1.25}\smash{\begin{tabular}[t]{l}car-like \\robot\end{tabular}}}}%
    \put(0.56556463,0.26641427){\makebox(0,0)[lt]{\lineheight{1.25}\smash{\begin{tabular}[t]{l}$x$\end{tabular}}}}%
    \put(0,0){\includegraphics[width=\unitlength,page=3]{framework_globalpathinvariant.pdf}}%
    \put(-0.00125021,0.15957524){\makebox(0,0)[lt]{\lineheight{1.25}\smash{\begin{tabular}[t]{l}$\mathcal{C}$\end{tabular}}}}%
    \put(0,0){\includegraphics[width=\unitlength,page=4]{framework_globalpathinvariant.pdf}}%
    \put(0.1328579,0.19967749){\makebox(0,0)[lt]{\lineheight{1.25}\smash{\begin{tabular}[t]{l}\textcolor{red}{motion planner}\end{tabular}}}}%
    \put(0,0){\includegraphics[width=\unitlength,page=5]{framework_globalpathinvariant.pdf}}%
    \put(0.68805988,0.04906196){\makebox(0,0)[lt]{\lineheight{1.25}\smash{\begin{tabular}[t]{l}switching scheme\end{tabular}}}}%
    \put(0,0){\includegraphics[width=\unitlength,page=6]{framework_globalpathinvariant.pdf}}%
    \put(0.61953684,0.05259325){\makebox(0,0)[lt]{\lineheight{1.25}\smash{\begin{tabular}[t]{l}$\mathcal{C}$\end{tabular}}}}%
    \put(0,0){\includegraphics[width=\unitlength,page=7]{framework_globalpathinvariant.pdf}}%
    \put(0.63517325,0.00332311){\makebox(0,0)[lt]{\lineheight{1.25}\smash{\begin{tabular}[t]{l}$y$\end{tabular}}}}%
    \put(0,0){\includegraphics[width=\unitlength,page=8]{framework_globalpathinvariant.pdf}}%
    \put(0.30624718,0.2209551){\makebox(0,0)[lt]{\lineheight{1.25}\smash{\begin{tabular}[t]{l}{\color{red} motion plan}\end{tabular}}}}%
    \put(0,0){\includegraphics[width=\unitlength,page=9]{framework_globalpathinvariant.pdf}}%
    \put(0.11902631,0.14665981){\makebox(0,0)[lt]{\lineheight{1.25}\smash{\begin{tabular}[t]{l}\textcolor{red}{global controller}\end{tabular}}}}%
    \put(0,0){\includegraphics[width=\unitlength,page=10]{framework_globalpathinvariant.pdf}}%
    \put(0.11868409,0.03356299){\makebox(0,0)[lt]{\lineheight{1.25}\smash{\begin{tabular}[t]{l}\textcolor{blue}{local controller}\end{tabular}}}}%
    \put(0,0){\includegraphics[width=\unitlength,page=11]{framework_globalpathinvariant.pdf}}%
  \end{picture}%
\endgroup%

%% file: tex/control.tex
\section{Hybrid Control Framework} \label{sec:control}
The convergence of $\xi$ and $\eta$ states to the desired set is valid only when the initial position of the robot is within \pn{$\mc{N}_\Gamma^{\by}$}. To guarantee the global convergence and path invariance, this paper proposes a strategy that generates a motion plan from the initial state to the desired path and employs a global tracking controller $\kappa_1:\Real^4 \to \Real^2$ to track the generated motion plan. As a result, the robot enters the neighborhood of the desired path within a finite time. Through a robust uniting control framework in \cite{San2021}, the local path-invariant controller $\kappa_0$ is activated, leveraging its convergence and invariance properties to ensure global convergence and path invariance.

\subsection{Motion Plan Generation}
\label{sec:trajectory_generation}
The foremost step in this strategy is to generate a motion plan from the current position to the path. This employs the motion planning technique to solve the following motion planning problem for (\ref{eq:car_robot}): 
% To relieve the curse of dimensions in motion planning, a simplified model of~\eqref{eq:car_robot} with state $\Tilde{x}:= (x_{1}, x_{2}, x_{3 })$ is considered in the motion planning software as
% \begin{equation}\label{eq:simplified_model}
% \dot{\Tilde{x}} = \begin{bmatrix}
%     \dot{x}_{1}\\
%     \dot{x}_{2}\\
%     \dot{x}_{3}
% \end{bmatrix} = 
%     \begin{bmatrix}
%         v\cos{x_{3}}\\
%         v\sin{x_{3}}\\
%         \frac{v\tan{\delta}}{l},
%     \end{bmatrix},
% \end{equation}
% where the velocity $v\in [v_{min}, v_{max}]$ is considered as a constant parameter and the steering angle $\delta\in [\delta_{min}, \delta_{max}]$ is considered as an input.
\begin{problem}\label{problem:mp}
    Given the initial state of the robot $x_{0}\in \mathbb{R}^{4}$, the final state set $X_{f} := \{(x_{1}, x_{2}, x_{3}, x_{4})\in \mathbb{R}^{4}: \exists (x_{5}, x_{6})\in \mathbb{R}^{2} \text{ such that } (x_{1}, x_{2}, x_{3}, x_{4}, x_{5}, x_{6})\in \Gamma\}$, the arbitrary unsafe set $X_{u}\subset \mathbb{R}^{4}$ that denotes the obstacles in the simulation \myifconf{as in Figures \ref{fig:sim}}{and experiments as in Figures \ref{fig:sim} and \ref{fig:exp1}}, and the system model (\ref{eq:car_robot}), the motion planning module generates a motion plan $(x'_{1}, x'_{2}, x_{3}', x_{4}'):[0, T]\to \mathbb{R}^{4}$ for some $T > 0$ such that\myifconf{1) $(x'_{1}(0), x'_{2}(0), x_{3}'(0), x'_{4}(0)) = x_{0}$;
    2) $(x'_{1}(T), x'_{2}(T), x_{3}'(T), x'_{4}(T))\in X_{f}$;
    3) there exists an input trajectory $(v', \omega') :[0, T]\to \mathbb{R}^{2}$ such that the state trajectory $(x'_{1}, x'_{2}, x_{3}', x_{4}')$ with input trajectory $(v', \omega')$ satisfies (\ref{eq:car_robot});
    4) there does not exist $t\in [0, T]$ such that $(x'_{1}(t), x'_{2}(t), x_{3}'(t), x_{4}'(t))\in X_{u}$.}{\begin{enumerate}
    \item $(x'_{1}(0), x'_{2}(0), x_{3}'(0), x'_{4}(0)) = x_{0}$;
    \item $(x'_{1}(T), x'_{2}(T), x_{3}'(T), x'_{4}(T))\in X_{f}$;
    \item there exists an input trajectory $(v', \omega') :[0, T]\to \mathbb{R}^{2}$ such that the state trajectory $(x'_{1}, x'_{2}, x_{3}', x_{4}')$ with input trajectory $(v', \omega')$ satisfies (\ref{eq:car_robot});
    \item there does not exist $t\in [0, T]$ such that $(x'_{1}(t), x'_{2}(t), x_{3}'(t), x_{4}'(t))\in X_{u}$.
\end{enumerate}}

\end{problem}
\myifconf{If no solution to Problem \ref{problem:mp} exists, the desired path $\Gamma$ is unreachable from the given initial state, making it impossible to guide the robot toward $\Gamma$. To the best of the authors' knowledge, no theoretical results currently verify the existence of a motion plan. Assuming at least one exists, complete motion planners are guaranteed to find it, though they are challenging to implement in practice. This paper uses the HyRRT motion planning tool from \cite{wang2022rapidly, wang2024motion, wang2023hysst, xu2024chyrrt, wang2024hyrrt, wang2023hysst1}, which is probabilistically complete and suitable for systems like (\ref{eq:car_robot}), despite being designed for hybrid systems.}{If no solution to Problem \ref{problem:mp} exists, then the desired path $\Gamma$ is not reachable from the given initial state, and, hence, it is impossible to drive the robot toward $\Gamma$. From the authors' best knowledge, there are no existing theoretical results to verify the existence of the motion plan. Assuming that at least one motion plan exists, complete motion planners are guaranteed to find it. However, in practice, the complete motion planner is difficult, if not impossible, to implement. In this paper, the HyRRT motion planning software tool in \cite{wang2022rapidly} is probabilistically complete and, though designed for hybrid systems, is suitable to generate the motion plan for systems like (\ref{eq:car_robot}).}
% Wang’s algorithm uses the hybrid model of the system to propagate forward in time from the initial position and backward in time from the target set. The two propagations are concatenated when an overlap is found to form the trajectory. It takes as inputs a hybrid system, a target set $T \subset \mathbb{R}^3$, and an unsafe set $U \subset \mathbb{R}^2$. The unsafe set is important because it allows our trajectory to consider obstacles when generating the trajectory. 
Since the motion plan is collision-free, the proposed hybrid controller is able to avoid the obstacles outside $\mathcal{N}_\Gamma^{\by}$. 
% This is illustrated in Figure \ref{fig:trajectory}, with the red boxes indicating obstacles to be avoided. 
% The offset of the generated trajectory to the target path in Figure \ref{fig:circletrajectory} is due to how the target set is defined. Since we know that $\kappa_0$ can converge to the path when inside the neighborhood, the trajectory generation does not have to end exactly on the path. This allows more room for other parameters to be factored in, such as the steepness of turning or the angle at which the robot arrives. Hence, the final position of the generated trajectory is a perfectly suitable initial position for $\kappa_0$.

% \begin{figure}[th]
%     \centering
%     \includegraphics[width = \columnwidth]{Figures/aux_traj.eps}
%     \caption{motion plan generated by motion planning module.}
%     \label{fig:trajectory}
% \end{figure}

% Also, note that our target set is a subset of $\mathbb{R}^3$. This is because the target set considers the $x$ and $y$ position and the orientation $\theta$. It could also consider $\delta$, but this is unnecessary for this paper’s purpose due to the nature of $\kappa_0$.

\subsection{Global Tracking Control and A Pure Pursuit Control Implementation} \label{sec:purepursuit}
A global tracking controller is employed as $\kappa_1$ to track the motion plan. To ensure that the global tracking controller effectively steers the car-like robot towards the motion plan and ultimately reaches the path's neighborhood, we impose the following assumption on $\kappa_{1}$.
\begin{assumption}\label{assumption:globalconvergence}
    Given a motion plan $x':[0, \infty)\to \mathbb{R}^{4}$, \nw{$x'$ is stable for the car-like robot controlled by $\kappa_1$,} namely, for all $\epsilon > 0$, there exists $\delta > 0$ such that $|\phi(t) - x'(t)| \leq \epsilon$
    for all $t \geq \nw{\delta}$, where $\phi:[0, \infty)\to \mathbb{R}^{4}$ is the maximal solution to (\ref{eq:car_robot}) with $(v,\omega) = \kappa_1(x, u)$.
\end{assumption}
\begin{remark}
    Assumption \ref{assumption:globalconvergence} ensures the car-like robot reaches the neighborhood of the desired path within a finite time. We choose $\epsilon = n_{c}$ in (\ref{eq:nbh_lift_set}). Since $x'$ is a solution to Problem \ref{problem:mp} and $x'(0) = \phi(0)$ (see item 1 in Problem \ref{problem:mp}), we have $|\phi(0) - x'(0)| = 0 \geq \delta$ for any existing $\delta > 0$ in Assumption~\ref{assumption:globalconvergence}. This implies $|\phi(t) - x'(t)| \geq \epsilon = n_{c}$ holds for all $t \geq 0$. By item 2 in Problem \ref{problem:mp}, there exists $T > 0$ such that $ x'(T)\in X_{f} = \{(x_{1}, x_{2}, x_{3}, x_{4})\in \mathbb{R}^{4}: \exists (x_{5}, x_{6})\in \mathbb{R}^{2} \text{ such that } (x_{1}, x_{2}, x_{3}, x_{4}, x_{5}, x_{6})\in \Gamma\}$. Hence, at time $T$, $|\phi(T) - x'(T)| < n_{c}$, implying the robot enters the neighborhood, namely $\phi(T)\in\mathcal{N}_\Gamma^{\by}$.
\end{remark}

Stability is a fundamental requirement in control design, and numerous tracking control techniques, such as pure pursuit control~\cite{Tomlin-PurePursuie-2011} and model predictive control~\cite{nascimento2018nonholonomic}, fulfill Assumption \ref{assumption:globalconvergence}. In this study, we employ the classic pure pursuit control as the global tracking controller for illustrative purposes. \myifconf{}{The pure pursuit algorithm calculates a steering angle that leads the robot on an arc path through a look-ahead point~\cite{Tomlin-PurePursuie-2011}.
%point some distance away on the path. 
This distance to the look-ahead point is called the look-ahead distance and can be tuned with a gain proportional to the robot’s speed. 
%
% Figure \ref{fig:purepursuit} gives a visual rendering of how the steering angle relates to the orientation of the robot and the angle to the path. The target point $(x_t, y_t)$ is found at a look-ahead distance $l_d$ away. The angle $\alpha$ is the difference between the robot's orientation and the angle to the target point. 
The look-ahead point $(x_t, y_t)\in \Real^{2}$ is found at a look-ahead distance $l_d\in\Real_{>0}$ away. The angle $\alpha_{p}$
% {\myred (AA: We have used $\alpha$ before in designing $\kappa_0$)} 
is the difference between the robot's orientation and the angle to the look-ahead point computed as
% \begin{equation} \label{eq:9}
$
    \alpha_{p} = x_{3} - \tan^{-1}{\left({(y_t - x_{2})}/{(x_t - x_{1})}\right)}.
$
% \end{equation}
The steering angle that leads the robot toward the look-ahead point is computed from $\alpha$ as
% \begin{equation} \label{eq:10}
$
    \delta = -\tan^{-1}\left( {(2l\sin{\alpha_{p})}}/{(l_d)}\right),
$
% \end{equation}
where $l$ is the length of the robot. The selection of the look-ahead point and the computation of the steering angle $\delta$ are executed in a receding manner to track the motion plan. \begin{remark}
    Only the position states, namely, $x_{1}$ and $x_{2}$, of the motion plan to Problem \ref{problem:mp} are used in the pure pursuit tracking algorithm.
\end{remark}}
%
% \begin{figure}
%     \centering
%     \includegraphics[width = 0.8\columnwidth]{Figures/simPurePursuit.eps}
%     \caption{Pure pursuit controller to track the generated motion plan.}
%     \label{fig:sin pure}
% \end{figure}
% The value of $\delta$ from Equation \eqref{eq:10} can be applied with sample and hold to drive our robot toward a point on the path. 
%
% The algorithm chooses a new point on the path towards which to navigate at each time step. More optimal look-ahead gain tuning results in lower tracking error and more desirable robot motion\cite{novelpurepursuit}.  
%
%
\myifconf{Figure~\ref{fig:hybrid} shows}{Figures~\ref{fig:hybrid} and \ref{fig:bestGP} show} that the pure pursuit algorithm is able to navigate the robot \nw{into} the neighborhood of the desired path \nw{by tracking the motion plan} while avoiding obstacles. 
%The pure pursuit algorithm is terminated when the robot enters the neighborhood of the desired path, as can be seen by the red line in the same figures. It can also be observed from these figures that the orientation at which the robot would arrive at the path is desirable. This results from the trajectory generator, since the target set includes a range of desirable robot orientations for each point. 
%
From \cite{ollero1995stability}, the pure pursuit controller is proved to satisfy Assumption \ref{assumption:globalconvergence}, thereby establishing the finite-time stability of $\kappa_1$ for $\mathcal{N}_\Gamma^{\by}$. 
% \begin{lemma}\label{lem:kappa1}
%     Given a motion plan $x':[0, \infty)\to \mathbb{R}^{4}$, then there exists a look-ahead distance $l_{d}$ such that the car-like robot controlled by pure pursuit algorithm $\kappa_1$ is stable, namely, for all $\epsilon > 0$, there exists $\delta > 0$ such that $|\phi(0) - x'(0)| \leq \delta$ implies $|\phi(t) - x'(t)| \leq \epsilon$
%     for all $t \geq 0$, where $\phi:[0, \infty)\to \mathbb{R}^{4}$ is the state trajectory of the robot under the control of $\kappa_1$.
% \end{lemma}
% \begin{remark}
%     Lemma \ref{lem:kappa1} ensures the car-like robot reaches the neighborhood of the desired path within a finite time. We choose $\epsilon = n_{c}$ in (\ref{eq:nbh_lift_set}). Since $x'$ is a solution to Problem \ref{problem:mp} and $x'(0) = \phi(0)$ (see item 1 in Problem \ref{problem:mp}), we have $|\phi(0) - x'(0)| = 0 < \delta$ for any existing $\delta > 0$ in Lemma \ref{lem:kappa1}. This implies $|\phi(t) - x'(t)| < \epsilon = n_{c}$ holds for all $t \geq 0$. By item 2 in Problem \ref{problem:mp}, there exists $T > 0$ such that $ x'(T)\in X_{f} = \{(x_{1}, x_{2}, x_{3}, x_{4})\in \mathbb{R}^{4}: \exists (x_{5}, x_{6})\in \mathbb{R}^{2} \text{ such that } (x_{1}, x_{2}, x_{3}, x_{4}, x_{5}, x_{6})\in \Gamma\}$. Hence, at time $T$, $|\phi(T) - x'(T)| < n_{c}$, implying the robot enters the neighborhood, namely $\phi(T)\in\mathcal{N}_\Gamma$.
% \end{remark}

\subsection{Hybrid Control Framework and Closed-loop System}
\begin{figure}[htbp]
    \centering
	\def\svgwidth{0.5\columnwidth}
	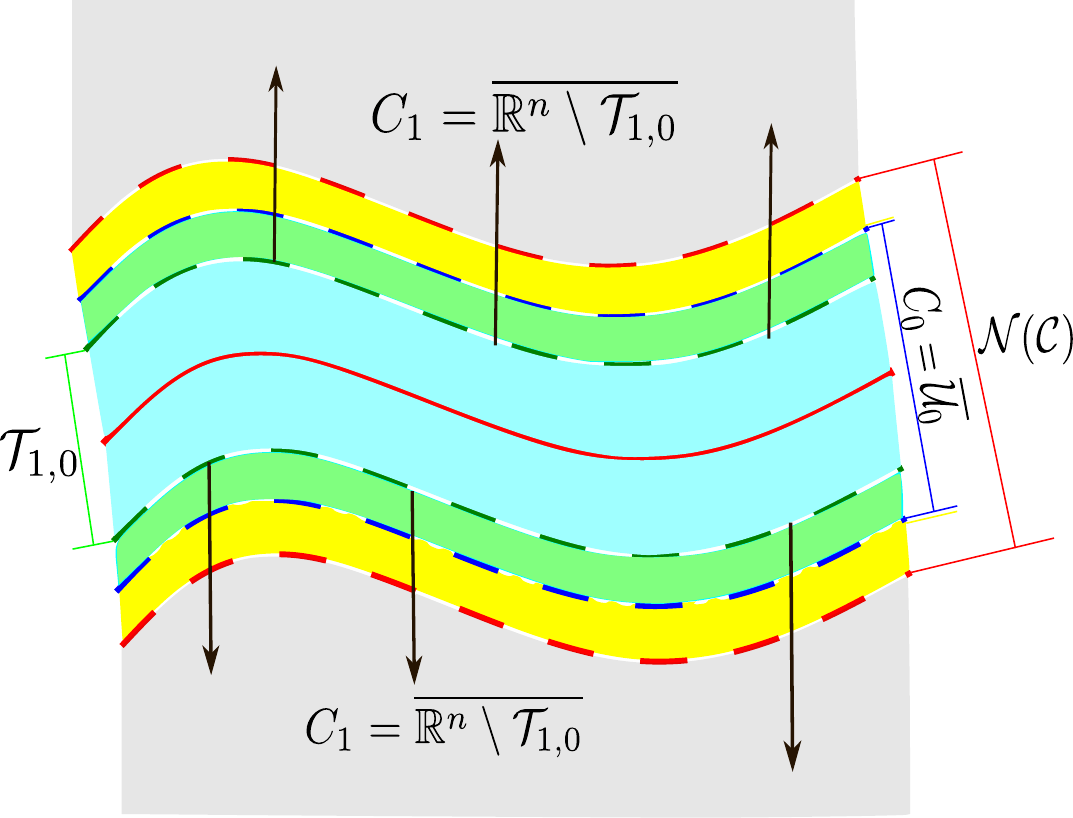

    \caption{\myifconf{The desired path $\Gamma$ is shown as a red solid line. The flow sets $C_{0}$ and $C_{1}$ are depicted in green and yellow, respectively, with their overlap also shown in green. Green dotted lines mark the boundaries of $C_{1}$, blue dotted lines indicate the boundaries of $C_{0}$, and red dotted lines represent the boundaries of $\mathcal{N}_{\Gamma}^{\by}$.}{The desired path $\Gamma$ is represented by the red solid line. The flow sets $C_{0}$ and $C_{1}$ are represented by the green region and yellow region, respectively, and the overlapped region between $C_{0}$ and $C_{1}$ are presented by the green region. The green dotted lines denote the boundaries of $C_{1}$ and the blue dotted lines denote the boundaries of $C_{0}$. The red dotted lines represent the boundaries of $\mathcal{N}_{\Gamma}^{\by}$.}}
    \label{fig:rough_fig}
    \vspace{-0.6cm}
\end{figure}
% The controller $\kappa_0$ renders the path invariant if the robot is initialized in the neighborhood of the path $\mc{N}_\Gamma$. 
A discontinuous, non-hybrid switching scheme could suffice for achieving global path invariance. However, this solution is sensitive to arbitrarily small noise and, therefore, is nonrobust. To overcome this issue, we design a hysteresis-based hybrid controller that is triggered by the distance to the path. 
% If the robot is initialized outside $\mc{N}_\Gamma$, the controller $\kappa_1$ forces the system to reach the neighborhood $\mc{N}_\Gamma$ in finite time.} 
For $0<c_1<c_{1,0}<c_0 < 1$, we can define the set $\mc{U}_{0}$ as follows:
\myifconf{$
\mc{U}_{0} \eqdef \set{\agx\in\Real^6 : \norm{\agx}_\Gamma < c_0 n_c  }, \;\; {\mc{U}_{0}} \subset \ak{\mc{N}_\Gamma^{\by}}.
$}{\[
\mc{U}_{0} \eqdef \set{\agx\in\Real^6 : \norm{\agx}_\Gamma < c_0 n_c  }, \;\; {\mc{U}_{0}} \subset \ak{\mc{N}_\Gamma^{\by}}.
\]}
Next, we define $\mc{T}_{1,0}$ such that $\mc{T}_{1,ff0}$ is contained in the interior of $\mc{U}_0$ as follows
\begin{equation}\label{eq:T10}
  \mc{T}_{1,0} \eqdef \set{\agx\in\Real^6 : \norm{\agx}_\Gamma {\leq} c_{1,0}n_c } \subset \mc{U}_{0}.  
\end{equation}
It is guaranteed by ~\cite[Proposition III.3]{AkhNieWas2015} that once the solution enters $\mc{T}_{1,0}$, it never reaches the boundary of $\overline{\mc{U}_0}$.
Let $C_{0} \eqdef \overline{\mc{U}_0}$ and $C_{1} \eqdef \overline{\Real^6\setminus\mc{T}_{1,0}}$, which lead to the hysteresis region $C_{0}\setminus \mc{T}_{1,0}$. The hybrid controller $\mc{H}_K = (C_{K}, F_{K}, D_{K}, G_{K})$ \mynne{takes the state $\agx \in \Real^6$ of (\ref{eq:dynamic_car_robot}) as its input and $q \in Q \eqdef \set{0,1}$ as its state, and can be modeled as in~(\ref{model:generalhybridsystem})}
%with state $q \in Q \eqdef \set{0,1}$, input $\agx \in \Real^6$ 
as follows:
\begin{subequations}\label{eq:Hyb-control}
\begin{align}
\label{eq:Hyb-control-1}
C_{K} &:= \bigcup_{q\in Q}\left( \set{q}\times  C_{K,q}  \right),\quad
%\end{equation}
%\begin{equation}
\begin{cases}
C_{K,0} \eqdef C_0\\ 
C_{K,1} \eqdef C_1\\
\end{cases}\\
% \end{equation}
% %
% \begin{equation}
\label{eq:Hyb-control-4}
    F_{K}(q, \agx) &:= 0\quad \forall (q,\agx) \in C_{K}\\
% \end{equation} 
%
% \begin{equation}
\label{eq:Hyb-control-2}
D_{K} &:= \bigcup_{q\in Q}\left( \set{q} \times D_{K,q}
\right),\quad
%\end{equation}
%\begin{equation}
\begin{cases}
D_{K,0} \eqdef \overline{\Real^6\setminus\mc{U}_{0}}\\ 
D_{K,1} \eqdef {\mc{T}_{1,0}}\\
\end{cases}\\
% \end{equation}
% %
% \begin{equation}
\label{eq:Hyb-control-5}
    G_{K}(q,\agx) &:= 1 -q \quad \forall (q,\agx) \in D_{K}
\end{align}
\end{subequations}
and the output function $\kappa: Q\times \mathbb{R}^{6} \to \reals^{2}$ is such that
\begin{equation}
\label{eq:Hyb-control-3}
\kappa(q,\agx) = q\kappa_{1}(\agx) + (1-q)\kappa_{0}(\agx),
\end{equation}
where the controller $\kappa_0$ is the locally path invariant controller defined in~\eqref{eq:kappa_0} and $\kappa_1$ is the pure pursuit controller. Hysteresis is created by sets $\mc{U}_{0}$ and $\mc{T}_{1,0}$.
% with the boundary of $\mc{U}_{0}$ and $\mc{T}_{1,0}$ being the outer and inner portion of the hysteresis region, respectively. 
Controlling the continuous-time plant~\eqref{eq:dynamic_car_robot} by the hybrid controller results in a hybrid closed-loop system with states $z = (\agx,q)$ and dynamics 
% resulting from controlling $\mc{H}_P$ with the hybrid controller $\mc{H}_K = (C_K,F_K,D_K,G_K,\kappa)$ changes according to 
$
    \dot \agx = F_P(z,\kappa(z,q)), \quad \dot q = 0
$
during flows, and at jumps, the state is updated according to 
$
    \agx^{+} = \agx,\quad q^{+} = 1-q.
$
Finally, the hybrid closed-loop system $\mc{H} = (C,F,D,G)$ with the state $z = (\agx,q) \in \Real^6 \times Q =: Z$ has data given as 
\begin{equation}
\label{eq:data-CLS-circle}
\begin{aligned}
    C &\eqdef \{(\agx,q) \in Z : (q,\agx) \in C_{K} \}\\
    F(z) &\eqdef \left[\begin{array}{c}
        F_P(\agx,\kappa(q, \agx))   \\
         0
    \end{array}\right]\;\; \forall z \in C\\
    D &\eqdef \{(\agx,q) \in Z : (q,\agx) \in D_{K} \}\\
    G(z) &\eqdef \left[\begin{array}{c}
         \agx   \\
         1-q
    \end{array}\right]\;\; \forall x \in D.
\end{aligned}
\end{equation}
% {\myblue where $C_P \eqdef \Real^6$.}
Next, we state the main result of our paper.
\begin{theorem}
\label{theo:geometric-hybrid-cricle}
Given a set $\Gamma$ and the continuous-time plant in ~\eqref{eq:car_robot}, suppose Assumptions~\ref{ass:implicit}, ~\ref{ass:SteeringAngle}, and~\ref{assumption:globalconvergence} hold. Let the hybrid controller $\mc{H}_K$ with data $(C_K,F_K,D_K,G_K,\kappa)$ defined in~\eqref{eq:Hyb-control} and~\eqref{eq:Hyb-control-3}. Then, the following hold:
%, and the closed-loop system $\mc{H} = (C,F,D,G)$ defined in~\eqref{eq:data-CLS-circle}.

\begin{enumerate}
    \item [{1)}] The closed-loop system $\mc{H} = (C,F,D,G)$ with data in~\eqref{eq:data-CLS-circle} satisfies the hybrid basic conditions\cite[Definition 2.18]{San2021};
    \item [{2)}]Every maximal solution to $\mc{H}$ from $C \cup D$ is complete and exhibits no more than two jumps; 
    \item [{3)}]The set 
$
    \mc{A} = \Gamma^\star \times \set{0}
$
    is global and robust finite-time stable for $\mc{H}$ in the sense of~\cite[Definition 3.16]{San2021} and is forward invariant.
    % {\blue Should I call the bullet points a1, a2, ..., or any better suggestion?}
\end{enumerate}

\end{theorem}
% \begin{proof}
%     The proof follows along the lines of~\cite[Theorem 4.6]{San2021}.
%     % , and is removed because of space limitations. 
% \end{proof}
\myifconf{\begin{proof}
    For a detailed proof, see \cite{wang2025hybrid}. A sketch of the proof is provided as follows: By (\ref{eq:T10}), the set $\mc{T}_{1,0}$ is closed, implying that $C_{K,0}$,$C_{K,1}$,$D_{K,0}$ and $D_{K,1}$ are also closed. Moreover, since $C_K$ and $D_K$ are finite union of $C_{K,0}$,$C_{K,1}$,$D_{K,0}$ and $D_{K,1}$, they are also closed. By (\ref{eq:Hyb-control-4}) and (\ref{eq:Hyb-control-5}), the maps $F_K$ and $G_K$ are continuous. Additionally, the pure-pursuit controller $\kappa_1$ and $\kappa_0$ in (\ref{eq:kappa_0}) are continuous, ensuring that the resulting closed-loop system $\mc{H}$ satisfies the hybrid basic conditions, which proves item~1.

    To prove the completeness of the maximal solutions to $\mc{H}$, we proceed by contradiction. Suppose there exists a maximal solution with the initial state $z(0,0)\in C\cup D$ that is not complete. From~\cite[Proposition 2.34]{San2021}, either item b or item c must hold. However, by Lemma~\ref{lemma:invariance}, the controller $\kappa_0$ assures finite-time stability of the desired path $\Gamma^\star$ everywhere in a neighborhood of $\Gamma^\star$, ruling out item b. Moreover, it can be shown that $G(D) \subset C \cup D$, hence, ruling out item c. Since the maximal solution is assumed to be unique, therefore, the solution $z$ is complete, establishing the contradiction.
    To prove that every solution exhibits no more than two jumps, we analyze the behaviors of the solutions with all the three possible initial conditions: i) $z(0,0) \in C_{K,1} \times \set{1}$; ii) $z(0,0)\in D_{K,1},  \times \set{1}$; iii) $z(0,0) \in C_{K,0} \times \set{0}$. In all three cases, Assumption \ref{assumption:globalconvergence} and Lemma \ref{lemma:invariance} ensure that every maximal solution has at most two jumps, which proves item~2.

    The attractivity of $\mc{T}_{1,0}$ in finite time is established by Assumption~\ref{assumption:globalconvergence}, while Lemma~\ref{lemma:invariance} implies that the set $\mc{A}$ is finite-time stable for $\mc{H}$, thereby establishing global finite-time stability. Finally, since the hybrid system satisfies the hybrid basic condition and $\mc{A}$ is compact, it follows from~\cite[Theorem 3.26]{San2021} that $\mc{A}$ is robust in the sense of~\cite[Definition 3.16]{San2021}, which proves item 3 and completes the proof.
\end{proof}}{
\begin{proof}
The right-hand side of~\eqref{eq:dynamic_car_robot} is a continuous function of $(\overline{x},u)$. By (\ref{eq:T10}), the set $\mc{T}_{1,0}$ is closed. Moreover, the sets $C_{K,0}$,$C_{K,1}$,$D_{K,0}$, and $D_{K,1}$ are also closed. This implies that $C_K$ and $D_K$ are also closed, as these sets are finite union of closed sets. The maps $F_K$ and $G_K$ are continuous by construction. Moreover, both the pure-pursuit controller $\kappa_1$ and the locally path-invariant controller $\kappa_0$ in (\ref{eq:kappa_0}) are continuous. Hence, the resulting closed-loop system $\mc{H}$ satisfies the hybrid basic conditions, which proves item~1.  

We prove the completeness of the maximal solutions to $\mc{H}$ by contradiction. Suppose there exists a maximal solution with the initial state $z(0,0)$ in the set $C\cup D$ that is not complete. Let $(T,J) = \sup \dom z$ and since by assumption $z$ is not complete $T + J < \infty$. From~\cite[Proposition 2.34]{San2021}, either item b or item c has to hold. {By Lemma~\ref{lemma:invariance}, the controller $\kappa_0$ assures finite-time stability of the desired path $\Gamma^\star$ everywhere in a neighborhood of $\Gamma^\star$. Hence, maximal solutions to the closed-loop system under the effect of $\kappa_0$ are bounded and complete. If the solution start from $C_0 \setminus \mc{T}_{1,0}$, it will eventually reach $\mc{T}_{1,0}$ under the control of $\kappa_1$. Hence the maximal solutions of the closed-loop system remain bounded and complete.} Under the control of $\kappa_1$, solutions reach a neighbourhood of $\Gamma^\star$, which is bounded. Hence, the solutions under $\kappa_1$ are bounded. Therefore, item b in~\cite[Proposition 2.34]{San2021} is ruled out. It can be shown that $G(D) \subset C \cup D$, hence by item 3 in~\cite[Proposition 2.34]{San2021}, item c is also ruled out. Therefore, every maximal solution to $\mc{H}$ from $C\cup D$ is complete. 

To show that every solution exhibits no more than two jumps, it should be noted that for every solution $z$ to $\mc{H}$, $z(0,0) \in C \cup D$, and only the following three cases are possible:

\begin{enumerate}
    \item~\label{list:1} By Assumption \ref{assumption:globalconvergence}, if $z(0,0) \in C_{K,1} \times \set{1}$, the solution $z$ reaches the set $D_{K,1}$ in finite hybrid time as the plant states reaches $\mc{T}_{1,0}$. After a jump, the solution $z$ remains flowing in $\left( C_{K,0}\setminus D_{K,0} \times \set{0}  \right)$ for all future hybrid time.
%     %{\blue I don't think we need to invoke the discussion of the set $\mc{E}_0$?}
%     %
    \item The solution exhibits the same behavior as in item~\ref{list:1}, when $z(0,0)\in D_{K,1} \times \set{1}$.
    \item If $z(0,0) \in C_{K,0} \times \set{0}$, then the following two cases are possible. If $z(0,0) \in \mc{T}_{1,0} \times \set{0}$, then by Lemma~\ref{lemma:invariance}, the solution remains flowing in $\left(  C_{K,0} \cap {C}_0 \times \set{0}\right)$ for all future hybrid time. If $z(0,0) \in \left( C_{K,0} \setminus \mc{T}_{1,0} \times \set{0}  \right)$, then the solution may either flow forever or jump from $0$ to $1$ when it reaches the boundary of ${C}_0$. From there, the solution flows according to the logic explained in item~\ref{list:1}.
\end{enumerate}
Hence, every maximal solution has at most two jumps, which proves item~2.

The attractivity of $\mc{T}_{1,0}$ in finite time is established by Assumption~\ref{assumption:globalconvergence}, while Lemma~\ref{lemma:invariance} implies that the set $\mc{A}$ is finite-time stable for $\mc{H}$, thereby establishing global convergence. Finally, since the hybrid system satisfies the hybrid basic condition and $\mc{A}$ is compact, it follows from~\cite[Theorem 3.26]{San2021} that $\mc{A}$ is robust in the sense of~\cite[Definition 3.16]{San2021}, which proves item 3 and completes the proof.
\end{proof}}

\subsection{Algorithm Formulation}
The hybrid controller switches between two controllers, defined by $\kappa_q$, with the hybrid model governing the state of $q$. This switching leverages each controller’s strengths based on the robot’s state. $\kappa_1$ guides the robot to the desired path’s neighborhood, while $\kappa_0$ ensures path tracking and invariance within it. The path-following scheme, guaranteeing invariance and global convergence, is detailed in Algorithm \ref{alg:globallyinvariant}.
% The hybrid controller switches between two controllers defined by the function $\kappa_q$, where our hybrid model governs the state of $q$. 
% Switching between the two controllers allows us to take advantage of the different assets depending on the robot’s state. 
% The controller $\kappa_1$ will navigate the robot to the neighborhood of the desired path if the robot starts or slides outside it. The controller $\kappa_0$ will be used inside the neighborhood to track the path and make it invariant for the closed-loop system. 
% The path-following scheme that renders invariance and guarantees convergence from everywhere in the output space is formulated in Algorithm \ref{alg:globallyinvariant}. 
{\footnotesize \begin{algorithm}[htbp]
    \caption{\footnotesize Hybrid globally path-invariant algorithm}\label{alg:globallyinvariant}
    \hspace*{\algorithmicindent} \textbf{Input:} The initial state $\agx_{0}$ of the robot.
    \footnotesize
\begin{algorithmic}[1]
\State $q\leftarrow 0$.
\While{true}
\If {$\agx_{0}\in \mathcal{T}_{1, 0}$ or ($\agx_{0}\in \overline{\mc{U}}_{0}\backslash \mathcal{T}_{1, 0}$ and $q = 0$)}
\State $q \leftarrow 0$.
\While{$\agx(t)\in \overline{\mc{U}_0}$}
\State Apply $\kappa_{0}$ to track $\mathcal{C}$.
\EndWhile
\Else
\State $q\leftarrow 1$.
\State Compute an auxiliary collision-free trajectory $x'$ connecting $x_0$ and $X_{f}$ using motion planner.
\While{$\agx(t)\notin \mathcal{T}_{1, 0}$}
\State Apply $\kappa_1$ to track $x'$.
\EndWhile
\EndIf
\State $\agx_{0}\leftarrow \agx(t)$.
\EndWhile
\end{algorithmic}
\end{algorithm}}
% \subsection{Main Result}

%%%%%% Proof excluded from the camera ready version
{
}

%% file: Figures/pdfs/neiborhood2.pdf_tex
%% Creator: Inkscape inkscape 0.92.5, www.inkscape.org
%% Accompanies image file 'neiborhood2.pdf' (pdf, eps, ps)
\begingroup%
  \makeatletter%
  \providecommand\color[2][]{%
    \errmessage{(Inkscape) Color is used for the text in Inkscape, but the package 'color.sty' is not loaded}%
    \renewcommand\color[2][]{}%
  }%
  \providecommand\transparent[1]{%
    \errmessage{(Inkscape) Transparency is used (non-zero) for the text in Inkscape, but the package 'transparent.sty' is not loaded}%
    \renewcommand\transparent[1]{}%
  }%
  \providecommand\rotatebox[2]{#2}%
  \newcommand*\fsize{\dimexpr\f@size pt\relax}%
  \newcommand*\lineheight[1]{\fontsize{\fsize}{#1\fsize}\selectfont}%
  \ifx\svgwidth\undefined%
    \setlength{\unitlength}{514.93201017bp}%
    \ifx\svgscale\undefined%
      \relax%
    \else%
      \setlength{\unitlength}{\unitlength * \real{\svgscale}}%
    \fi%
  \else%
    \setlength{\unitlength}{\svgwidth}%
  \fi%
  \global\let\svgwidth\undefined%
  \global\let\svgscale\undefined%
  \makeatother%
  \begin{picture}(1,0.76224906)%
    \lineheight{1}%
    \setlength\tabcolsep{0pt}%
    \put(0,0){\includegraphics[width=\unitlength,page=1]{neiborhood2.pdf}}%
    \put(0.46022825,0.36474439){\color[rgb]{0,0,0}\makebox(0,0)[lt]{\lineheight{1.25}\smash{\begin{tabular}[t]{l}$\Gamma$\end{tabular}}}}%
  \end{picture}%
\endgroup%

%% file: tex/results.tex
\section{Simulation Results}\label{section:simulation}
The proposed algorithm is simulated on the MATLAB software\footnote{Code at \hyperlink{https://github.com/HybridSystemsLab/HybridGloballyPathInvariantControl.git}{https://github.com/HybridSystemsLab/PathInvariantControl.git}.}. 
\subsection{\nw{CBF-based Singularity Filter Prevents Singularity Points}}
\begin{figure}[t]
 \centering
 \begin{subfigure}{0.45\columnwidth}
    \centering
    \includegraphics[width = \columnwidth]{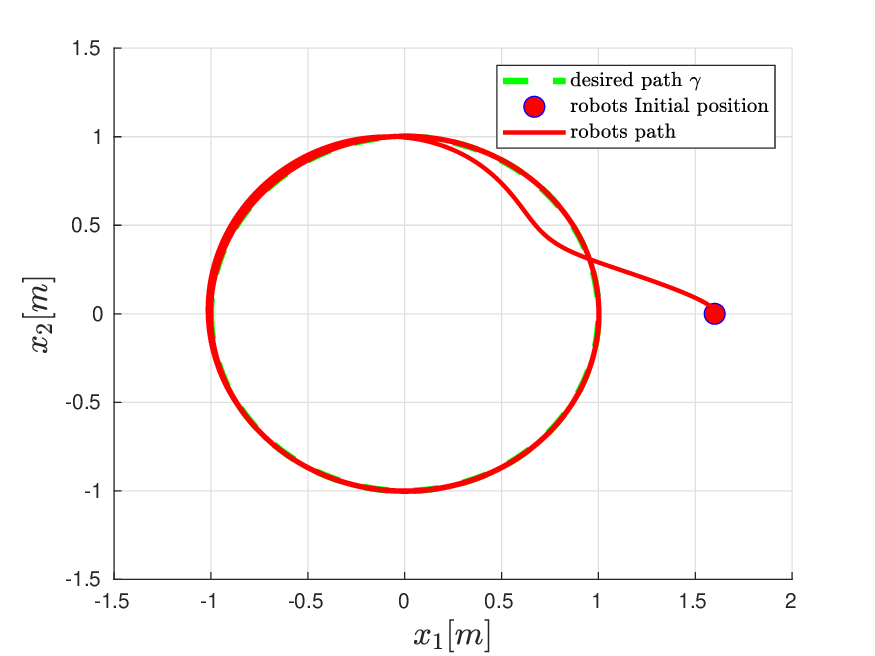}
    \caption{The controller $\kappa_{\xi}$ ensures finite-time convergence to the desired curve.}
    \label{fig:local-xy-position}
 \end{subfigure}
 \begin{subfigure}{0.45\columnwidth}
    \centering
    \includegraphics[width = \columnwidth]{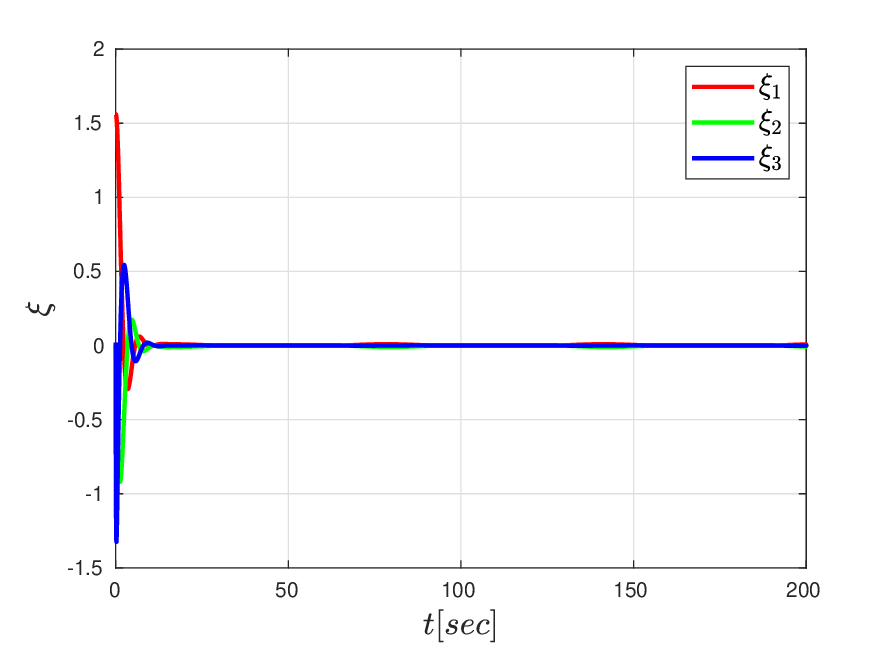}
    \caption{The controller $\kappa_{\xi}$ ensures the path is forward invariant.}
    \label{fig:local-xi-states}
 \end{subfigure}
 % \caption{Barrier certified local path invariant controller}
 \label{fig:sim}
 % \end{figure}
 % \begin{figure}[htbp]
 \centering
 \begin{subfigure}{0.45\columnwidth}
    \centering
    \includegraphics[width = \columnwidth]{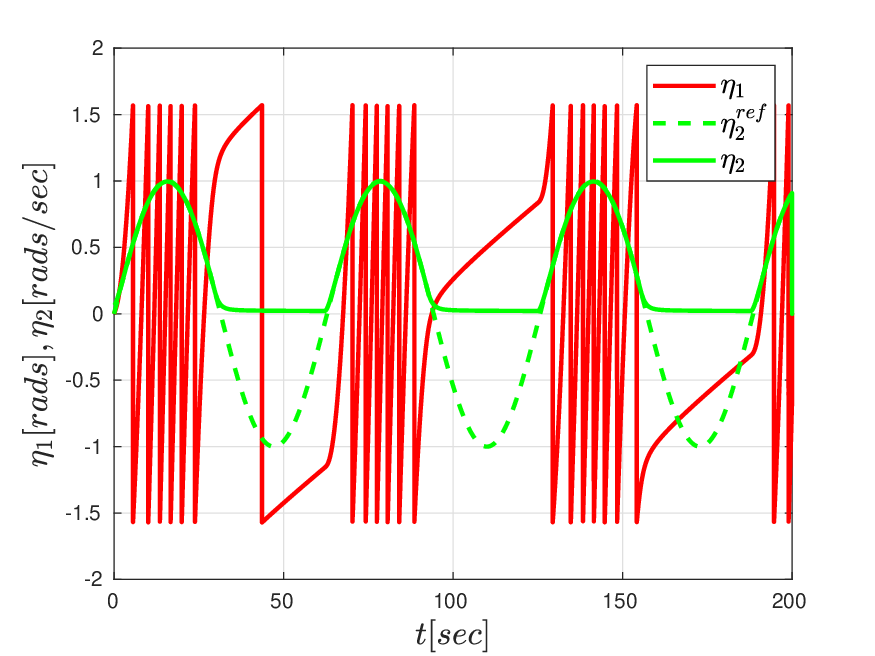}
    \caption{The barrier condition ensures that $\eta_2$ remains strictly positive.}
    \label{fig:local-eta1-eta2}
 \end{subfigure}
 \begin{subfigure}{0.45\columnwidth}
    \centering
    \includegraphics[width = \columnwidth]{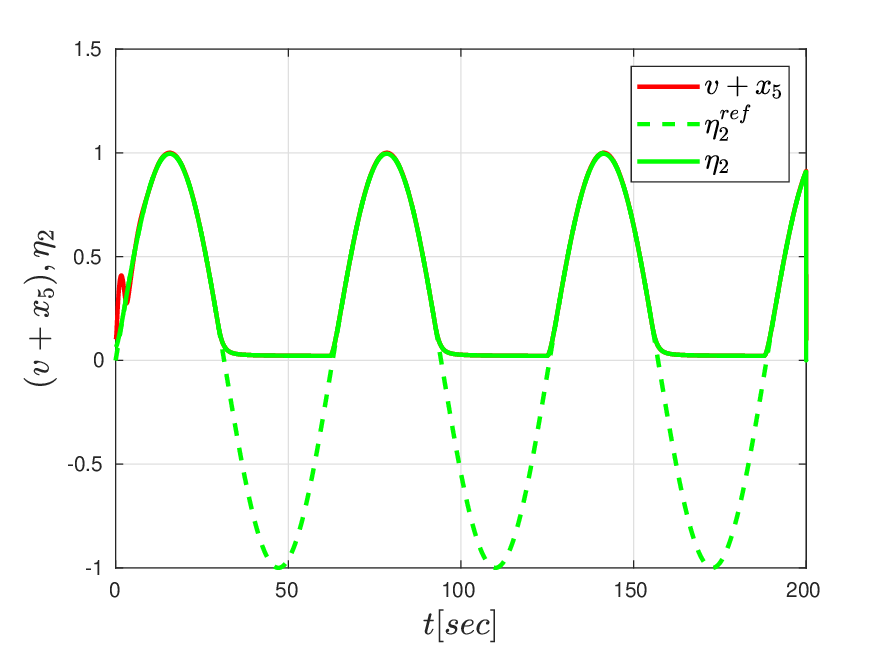}
    \caption{On the path, the forward velocity $\vrm + x_5 = \eta_2$.}
    \label{fig:local-speed}
 \end{subfigure}
 \caption{CBF-based singularity filter prevents singularity point.}
 \label{fig:sim}
  \vspace{-0.2cm}
 \end{figure}

We simulate the controller $\kappa_0 = (\kappa_\xi,\kappa_\eta)$ from~\eqref{eq:kappa_0}. To achieve finite-time path stability and invariance, we use $v^\pitchfork = \kappa_\xi(\xi)$ from~\eqref{eq:v_trans}. Figure~\ref{fig:local-xy-position} shows the system output converging to the desired path, while Figure~\ref{fig:local-xi-states} confirms $\xi$ remains on the path, ensuring forward invariance. To track the speed profile $\eta^{\mathrm{ref}}_2 = \sin(t)$ while satisfying the barrier constraint, an optimization-based control scheme is used.
% As shown in Figure~\ref{fig:local-xy-position}, the output of the system converges to the desired path, and as seen in Figure~\ref{fig:local-xi-states}, the $\xi$ state always remains on the path, i.e., the path is forwarded invariant. To track a given speed profile $\eta^{\mathrm{ref}}_2 = \sin(t)$ while ensuring the barrier constraint an optimization-based control scheme is proposed. 
Concretely, we solve the following Quadratic Program (QP) based controller:
\myifconf{$$
\begin{aligned}
\hspace{10pt} \arg \min_{(v^\parallel,\delta_s)} & \quad \frac{1}{2}(v^\parallel)^2 + p \delta_s^2\\
    % \textrm{s.t.} \;\quad \; \; Av  \leq &  \; b, \label{C1 cont const}\\
    \textrm{s.t.} \; L_{\tilde f}V(\eta) + & L_{\tilde g} V(\eta)v^\parallel \leq -\alpha_k (V(\eta)) + \delta_s\label{eq:QP_lyapunov}\\
    \; L_{\tilde f}^2b(\eta) + & L_{\tilde f}b(\eta) +L_{\tilde g}L_{\tilde f}b(\eta)v^{\parallel} 
    \leq -\alpha_k( b(\eta) + L_{\tilde f}b(\eta) ),\label{eq:QP_barrier}
\end{aligned}
$$}{\begin{subequations}\label{QP-based-control}
\begin{align}
\hspace{10pt} \arg \min_{(v^\parallel,\delta_s)} & \quad \frac{1}{2}(v^\parallel)^2 + p \delta_s^2\\
    % \textrm{s.t.} \;\quad \; \; Av  \leq &  \; b, \label{C1 cont const}\\
    \textrm{s.t.} \; L_{\tilde f}V(\eta) + & L_{\tilde g} V(\eta)v^\parallel \leq -\alpha_k (V(\eta)) + \delta_s\label{eq:QP_lyapunov}\\
    \; L_{\tilde f}^2b(\eta) + & L_{\tilde f}b(\eta) +L_{\tilde g}L_{\tilde f}b(\eta)v^{\parallel} 
    \leq -\alpha_k( b(\eta) + L_{\tilde f}b(\eta) ),\label{eq:QP_barrier}
\end{align}
\end{subequations}}
where $\delta_s$ is a relaxation variable that ensures the solvability of the QP as penalized by $p>0$. \myifconf{Figure~\ref{fig:local-eta1-eta2} shows that the barrier condition is never violated. The robot follows the reference velocity profile when $\eta_2 \geq 0.02$, and maintains a positive velocity when $\eta_2 < 0.02$. This guarantees forward invariance, as trajectories never reach $\vrm + x_5 =0$.}{As shown in Figure~\ref{fig:local-eta1-eta2}, the barrier condition is never violated. In other words, the robot follows the given reference velocity profile whenever $\eta_2 \geq 0.02$, and when $\eta_2 < 0.02$, the velocity of the robot stays at some positive distance away from zero. Therefore, \nw{forward invariance} is guaranteed, as the system trajectories never reach the singularity point $\vrm + x_5 = 0$.}
 \begin{figure}[t]
 \centering
 \begin{subfigure}{0.45\columnwidth}
    \centering
    \includegraphics[width = \columnwidth]{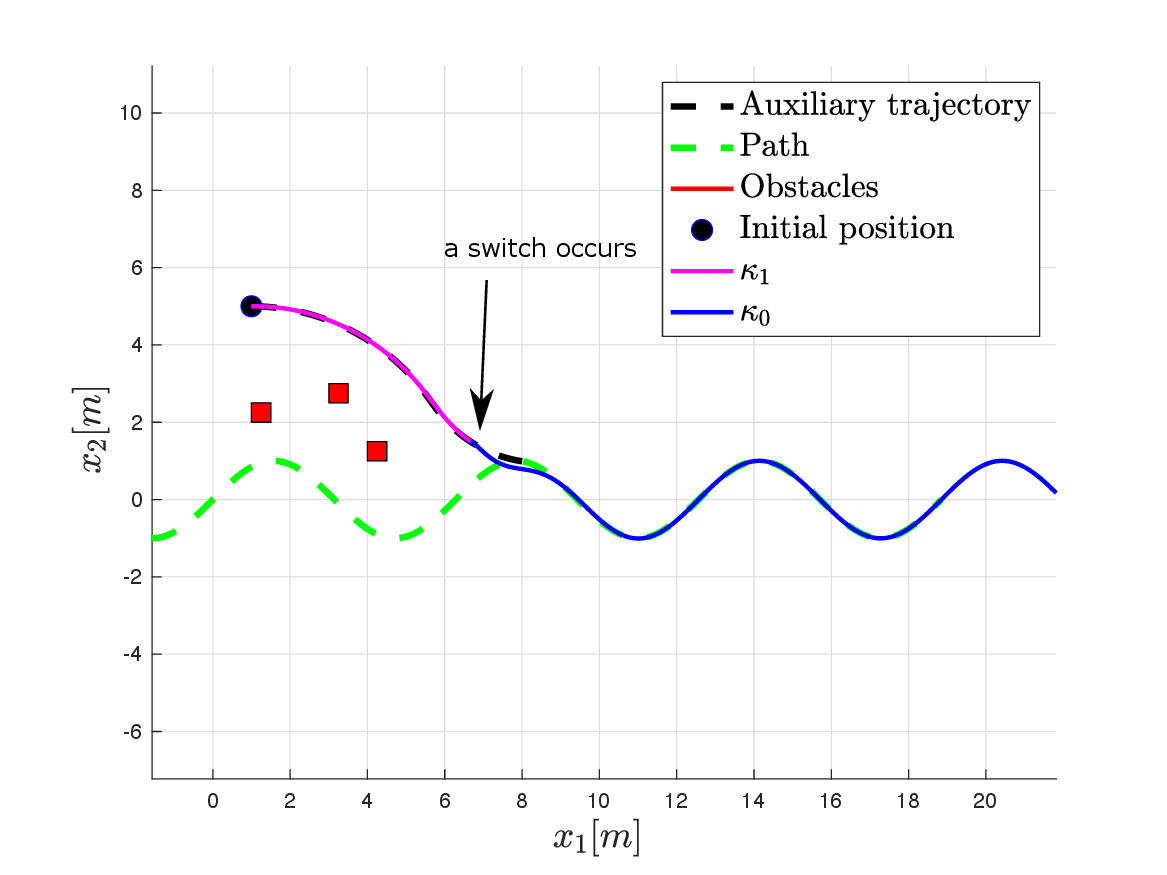}
    \caption{The simulated trajectory of the robot tracking a sinusoidal desired path.}
    \label{fig:hybrid}
 \end{subfigure}
 \begin{subfigure}{0.45\columnwidth}
    \centering
    \includegraphics[width = \columnwidth]{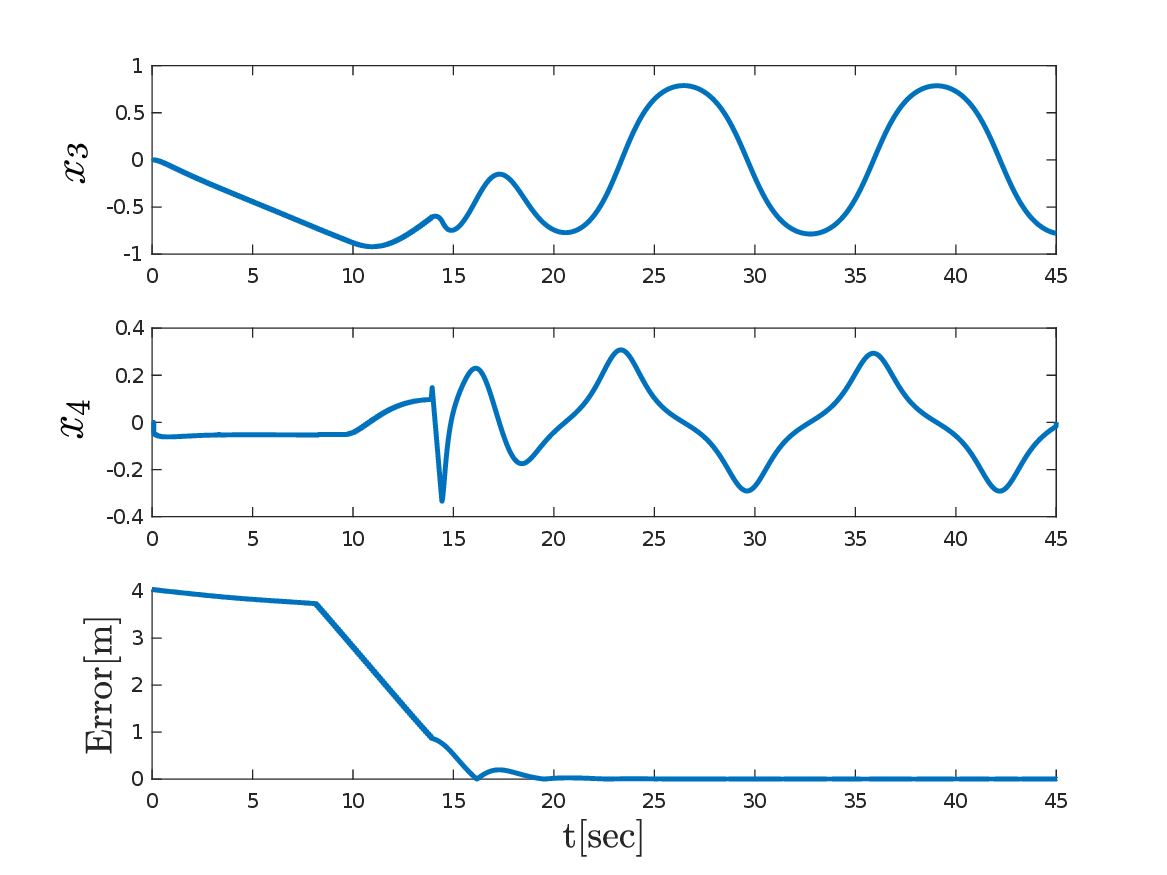}
    \caption{The evolution of $x_{3}$, $x_{4}$, and error. }
    \label{fig:states}
 \end{subfigure}
 \caption{Simulation results of the proposed hybrid controller.}
 \label{fig:sim}
 \vspace{-0.8cm}
 \end{figure}
\subsection{\nw{Hybrid Control Framework Achieves Global Path Invariance}}
\myifconf{We present hybrid control results that ensure path invariance with safety guarantees and finite-time global convergence. Figure \ref{fig:sim} shows the simulations. In Figure \ref{fig:hybrid}, the motion planner's auxiliary trajectory is a black dotted line, the desired sinusoidal path $\Gamma$ is green dotted, and obstacles are red squares. The robot's trajectory under $\kappa_{1}$ is purple (solid), and under $\kappa_{0}$ is blue (solid). The robot follows $\kappa_{1}$ until $t = 15$, then switches to $\kappa_{0}$, maintaining the path thereafter. Figure \ref{fig:states} shows the evolution of $x_{3}$, $x_{4}$, and the distance between the robot and the desired path. The bottom plot illustrates that the error decreases under $\kappa_{1}$ and remains zero after switching to $\kappa_{0}$, demonstrating path invariance.}{Next, we demonstrate the results of hybrid control, which not only renders the path invariant with a safety guarantee but also depicts global convergence to the given path in finite time. The simulation results are shown in Figure \ref{fig:sim}. In Figure \ref{fig:hybrid}, the auxiliary trajectory generated by the motion planning module is represented by the dotted line in black. The given sinusoidal desired path $\Gamma$ is represented by the dotted line in green. The obstacles are represented by the red squares. The solid line in purple denotes the trajectory of the robot under the control of $\kappa_{1}$ and the solid line in blue denotes the trajectory of the robot under the control of $\kappa_{0}$. In Figure \ref{fig:hybrid}, the robot is under the control of $\kappa_{1}$ before entering a neighborhood of $\Gamma$ at $t = 15$. Then the control is switched to $\kappa_{0}$ and the trajectory of the robot stays on the path afterward. Figure \ref{fig:states} shows the evolution of $x_{3}$, $x_{4}$, and the distance between the output position and desired path. As is shown in the bottom of Figure \ref{fig:states}, the error between the robot and the desired path decreases when the $\kappa_{1}$ is in control and the error remains zero after $\kappa_{0}$ takes over the control. This demonstrates the path-invariance property.}

% The error metric in this simulation is defined as follows
% $
%     \texttt{error}(t, x, q) := \left\{
%     \begin{aligned}
%         &e_{0}(x)&\text{ if } q = 0\\
%         &e_{1}(t, x)&\text{ if } q = 1
%     \end{aligned}
%     \right.
% $ where
% $
%     e_{0}(x) := |\sin(x_{1}) - x_{2}|
% $ and 
% $
%     e_{1}(t, x) := \sqrt{(x'_{1}(t) - x_{1}(t))^{2} + (x'_{2}(t) - x_{2}(t))^{2}}.
% $

\myifconf{}{\section{Experimental Results}\label{section:experiment}
\begin{figure}[htbp]
 \centering
 \begin{subfigure}{0.48\columnwidth}
    \includegraphics[width = \columnwidth]{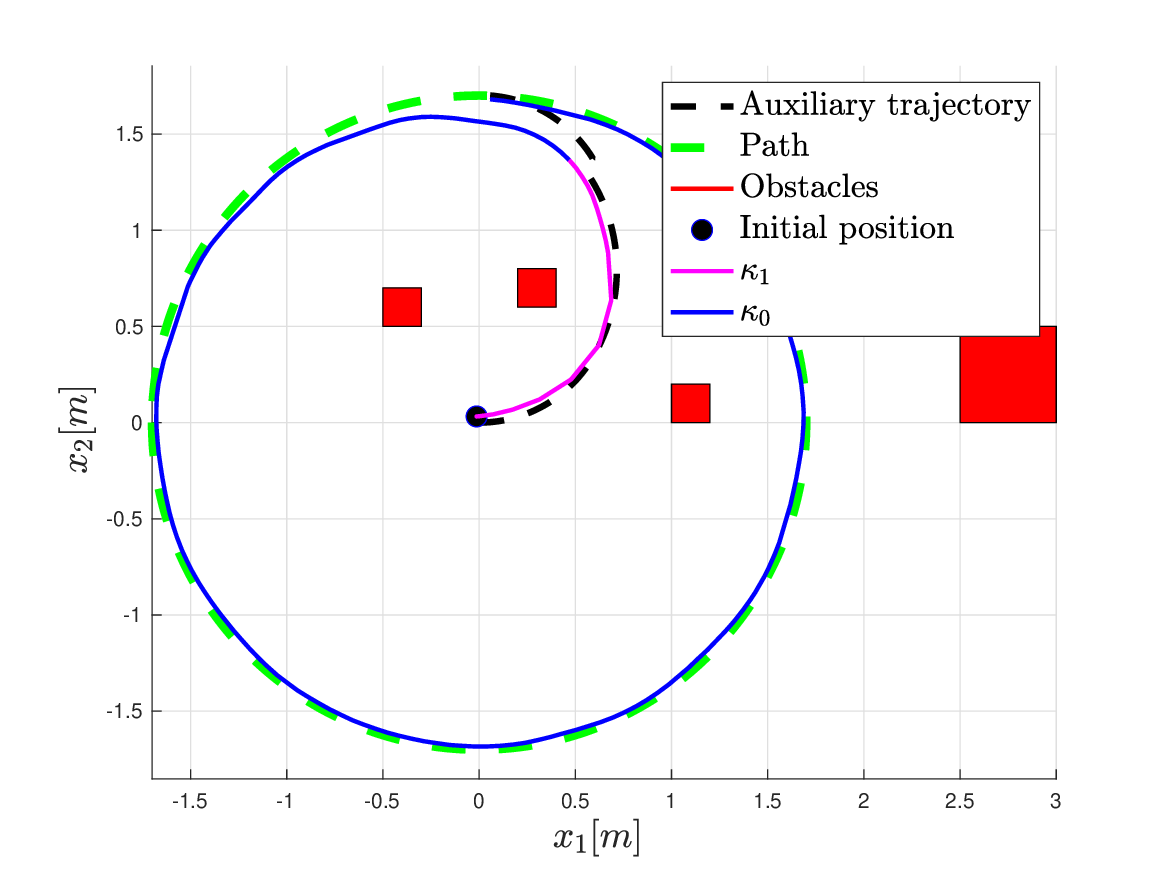}
    \caption[size=0.5]{Experiment result starting from the center of reference circle.}
    \label{fig:bestGP}
 \end{subfigure}
 \begin{subfigure}{0.48\columnwidth}
    \includegraphics[width = \columnwidth]{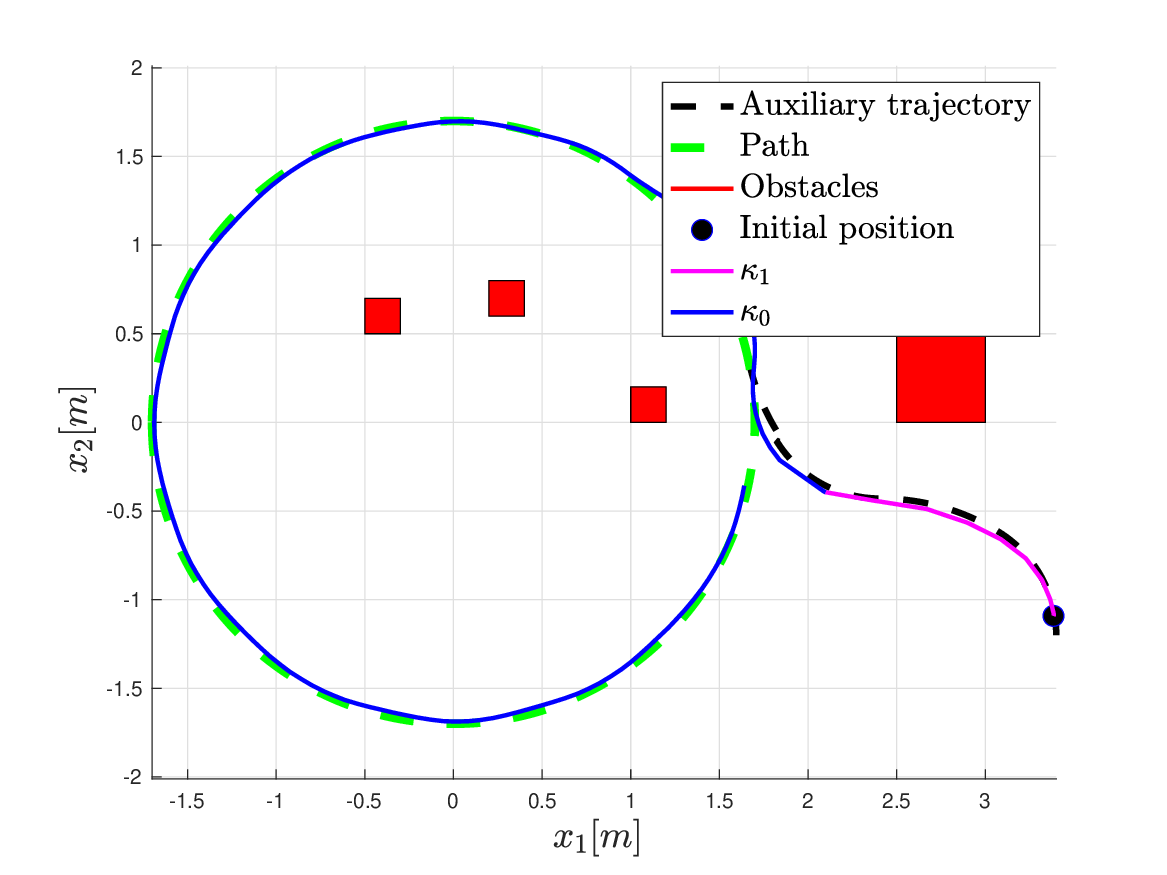}
    \caption{Experiment result starting from the outside of reference circle.}
    \label{fig:exp2}
\end{subfigure}
%  \begin{subfigure}{0.48\columnwidth}
%     \includegraphics[width = \columnwidth]{Figures/exp1Error.eps}
%     \caption{Distance error over time}
%     \label{fig:circleerror}
% \end{subfigure}
\caption{Experiment Results}
\label{fig:exp1}
\end{figure}
% In this experiment we use OSOYOO Robot Car that features a servo power steering motor that controls the axles to turn the wheels. This electric power steering system enables a highly automated driving experience that imitates real-world full-sized rear-wheel robots. The car has two DC motors with a maximum speed of 100 RPM. The motor driver is an L298N dual H-bridge driver, which allows controlling the direction and speed of the motors. The car is powered by a $9$V battery. The car is approximately $225$mm$\times$$150$mm$\times$$75$mm and weighs about $650$g, including the battery. The diameter and thickness of the wheels are $65$mm and $26$ mm, respectively. The car is equipped with an Arduino Uno board (referred to as the onboard computer), including a Bluetooth module, which enables the user to send and receive data (control command and system state) via an off-board computer. In our experimental setup, infrared markers are attached to the robot, and using a mocap indoor positioning system, the position and orientation of the robot are determined at the off-board computer. The steering angle information is sent from the onboard computer to the off-board computer. The hybrid control algorithm is run off-board, and the control commands are sent via Bluetooth to the onboard computer.   
The proposed algorithm is validated through real-world experiments, accounting for the presence of noise and model uncertainties. In the experiment, we use the OSOYOO Robot Car, which simulates real rear-wheel robots via a servo power steering motor that controls the axles. 
% The car boasts two DC motors with a max speed of $100$ RPM, managed by an L298N dual H-bridge driver. 
The car measures approximately $225$mm $\times$ $150$mm $\times$ $75$mm.
% and weighs $650$g with the battery. 
% It's equipped with an Arduino Uno board with Bluetooth for data exchange. 
In the experimental setup, the robot's position and orientation are tracked by a motion capture indoor positioning system. Steering commands are transmitted via Bluetooth from an off-board computer that executes the hybrid control algorithm. 
%Due to space limit, more details and experiment results are presented in  \cite{wang2023hybrid}.

% In the setup, infrared markers and a mocap indoor positioning system track the robot's position and orientation, while steering commands are sent via Bluetooth from an off-board computer running the hybrid control algorithm.

The results of running the proposed algorithm on the scaled car are shown in Figure \ref{fig:exp1}, with the desired path being a circle. In the experiments, a scaled car is controlled by the proposed algorithm to track the same desired path starting from different initial positions. 
% The results of running the proposed algorithm on the scaled car are shown in Figure \ref{fig:exp1}, with the reference path being a circle. In the experiments, a scaled car is controlled by the proposed algorithm to track the circle reference path starting from its center, which is a singularity point of $\kappa_0$. In the experiments tracking a circle path with radius $R$, the error metric is defined as
% The error metric is defined as
% $\label{eq:errorcircle}
%     \texttt{error}(t, x, q) := \left\{
%     \begin{aligned}
%         &e_{0}(x)&\text{ if } q = 0\\
%         &e_{1}(t, x)&\text{ if } q = 1
%     \end{aligned}
%     \right.
% $ where
% $
%     e_{0}(x) := \norm{\sqrt{x_{1}^{2} + x_{2}^{2}} - R}
% $ and 
% $
%     e_{1}(t, x) := \sqrt{(x'_{1}(t) - x_{1}(t))^{2} + (x'_{2}(t) - x_{2}(t))^{2}}.
% $

In the first experiment, the initial position is the center of the reference circle which is a singularity point of $\kappa_0$. In Figure \ref{fig:bestGP}, it is shown that the proposed algorithm firstly generates an auxiliary trajectory and employs pure pursuit controller $\kappa_{1}$ to track the auxiliary trajectory to leave the singularity point and to approach the desired path. After the robot arrives in the neighborhood of the desired path, the locally path-invariant controller $\kappa_{0}$ is triggered to control the robot and the robot stays invariant on the desired path. 
% The error during this process computed by (\ref{eq:errorcircle}) is shown in Figure \ref{fig:circleerror}. 
% The error decreases at the beginning when the pure pursuit controller is in use, meaning that the car is approaching to the reference path and leaving the singularity point. After the switch is triggered and $\kappa_{0}$ takes over the control at $t = 4.3$ seconds, the error stays constantly close to zero, which means that the car stays on the reference path.

 % In Figure \ref{fig:bestGP}, it is shown that the proposed algorithm  firstly generates an auxiliary trajectory and employs pure pursuit controller $\kappa_{1}$ to track the auxiliary trajectory to leave the singularity point and to approach to the reference path. After the robot arrives in the neighborhood of the reference trajectory, the locally path-invariant controller $\kappa_{0}$ is triggered to control the robot reaches and stay invariant on the reference path. The error during this process computed by (\ref{eq:errorcircle}) is shown in Figure \ref{fig:circleerror}. The error decreases at the beginning when the pure pursuit controller is in use, meaning that the car is approaching to the reference path and leaving the singularity point. After the switch is triggered and $\kappa_{0}$ takes over the control at $t = 4.3$ seconds, the error stays constantly close to zero, which means that the car stays on the reference path.

 In the second experiment shown in Figure \ref{fig:exp2}, the settings are the same except that the initial position of the robot is changed to $(3.5, -1.2)$ which is not a singularity but rather more general. Following the same procedure, the proposed algorithm starts with generating an auxiliary trajectory and using $\kappa_{1}$ when the robot is away from the desired path. Then it switches to $\kappa_{0}$ and stays invariant on the path when it is in the neighborhood of the desired path.}

%% file: tex/conclusion.tex
\vspace{-0.2cm}
\section{Conclusion} \label{sec:conclusion}
\myifconf{This paper presents a safe hybrid control framework for car-like robots, ensuring robustness, global path convergence, and path invariance. It switches between a locally path-invariant controller and a globally convergent pure pursuit controller within a robust uniting control framework. Using motion planning, it avoids obstacles beyond a tight neighborhood. Simulations demonstrate global path invariance and safety. We have conducted experiments validating the hybrid control framework on the OSOYOO Robot Car and are currently implementing the CBF-based path-invariant controller.}{This paper proposes a safe hybrid control framework for car-like robots that guarantees robustness, global convergence to the given path, and path-invariance properties. 
% The existing controllers for AVs only guarantee path invariance when starting within a tight neighborhood of the reference path and can not avoid obstacles. 
The proposed hybrid controller switches between a locally path-invariant controller and a globally finite-time convergent pure pursuit controller following a robust uniting control framework. 
% When the car is initially outside of the neighborhood of the reference path, a collision-free auxiliary trajectory is generated connecting the initial position with the reference path and the pure pursuit controller controls the vehicle into the neighborhood in finite time. Then the local path-invariant controller takes over to ensure the robustness, stability, and path-invariance. 
By employing motion planning technology, the proposed hybrid control scheme is able to avoid collision with obstacles outside the tight neighborhood at the same time. \pn{The main result showcases the global path-invariance of the proposed hybrid control framework, thereby ensuring its safety.} Those results are demonstrated both by simulation and experiments.}
% This paper proposed a hybrid controller that makes a large class of desired paths invariant and attractive. These findings were demonstrated on a small car-like robot and gave evidence to the previous claim. 

% There are many exciting ways the work in this paper can be expanded. Implementing the Extended Kalman Filter mentioned in Section \ref{sec:results}  would improve the hybrid controller by providing smoother controls and decreasing steady-state oscillations. An interesting situation is when the unsafe set overlaps the desired path. Currently, the hybrid controller would not be able to deal with this situation. Finally, an exciting next step for this project would be implementing this algorithm on a car that estimates its state through onboard sensors rather than Optitrack. This would allow for testing in more exciting and practical environments.

%% file: tex/appendix.tex
% \appendix
\myifconf{}{\section{Appendix}
\label{section:appendix}

% \begin{definition}(robust stability~\cite{San2021})
% \label{def:robust-stability}
% { Given a hybrid closed-loop system $\mc{H}$, a nonempty closed set $\mc{A}\subset \ms{M}$ and an open set $\mc{U}\subset \ms{M}$ such that $\mc{A} \subset \mc{U}$, the set $\mc{A}$ is said to be robustly stable for $\mc{H}$ on $\mc{U}$ if for every proper indicator function $\varpi$ of $\mc{A}$ on $\mc{U}$, every function $\beta \in \mc{KL}$ such that 
% \[
% \varpi(x(t,j)) \leq \beta(\varpi(x(0,0)), t+j)\quad \forall (t,j)\in\dom x 
% \]
% for the solutions to $\mc{H}$ from $\mc{U}$, and every continuous function 
% $\rho^{*}:\ms{M} \to \Real_{\geq 0}$ that is positive on $\mc{U} \setminus \mc{A}$, the following holds: for each compact set $K \subset \mc{U}$ and each $\epsilon >0$, there exists $\delta^{*} >0$ such that for each solution {$x_{\rho}$} the perturbed system $\mc{H}_{\rho}$ with $\rho = \delta^{*}\rho^{*}$, starting from $x_{\rho}(0,0) \in K$ satisfies 
% \[
% \varpi(x_{\rho}(t,j)) \leq \beta(\varpi(x_{\rho}(0,0)), t+j)+ \epsilon \quad   \forall (t,j)\in\dom x_{\rho}.
% \]}
% \end{definition}

{\myblue Consider a time-invariant, finite-dimensional, deterministic control-affine system with $m$ inputs, $u:=[u_1 \cdots u_m]^{\top}\in \mathbb{R}^m$ and $p$ outputs, along with smooth maps $f:\mathbb{R}^n\rightarrow \mathbb{R}^n$, $g_{i}:\mathbb{R}^n\rightarrow \mathbb{R}^n$ and $h:\mathbb{R}^n\rightarrow \mathbb{R}^p$.
\begin{eqnarray}\label{eq:most_general_sys_mathprelim}
%\begin{split}
 \dot{x}=f(x)+ \sum^{m}_{i=1}g_{i}(x)u_{i} \eqdef f(x)+ g(x)u,
 % \end{split}
\end{eqnarray}
and consider a function,
\begin{eqnarray}\label{eq:most_general_output_appendex}
%\begin{split}
  y=h(x)=\left[
  \begin{array}{c}
      h_1 (x)\\
      \vdots\\
      h_p (x)\\
  \end{array} \right],\forall y \in \mathbb{R}^p,
 % \end{split}
\end{eqnarray}
which is the output of the system. 
% The relative degree is the key concept in solving the feedback linearization problem.
% \begin{definition}(Relative degree)
% Consider system~\eqref{eq:most_general_sys_mathprelim} with $u \in \mathbb{R}$ and with output function~\eqref{eq:most_general_output_appendex} with $m=p=1$ i.e., $y=h(x)$, $y \in \mathbb{R}$. The system has a relative degree of $r$ at a point $x_0$ if
% \begin{enumerate}
%   \item $L_g L_f ^k h(x)=0$, for all $x \in $ a neighborhood of $x_0$ and for all $k<r-1$,
%   \item $L_g L_f ^{r-1} h(x_0)\neq 0$.
% \end{enumerate}
% \end{definition}
% The relative degree of a single input single output (SISO) system in the number of times we need to differentiate the output before the control input appears. 
The vector relative degree can be defined for the multiple-input multiple-output (MIMO) systems.

\begin{definition}(Vector relative degree \cite{Sastry})
\label{def:vector_relative_deg}
Consider system \eqref{eq:most_general_sys_mathprelim} with $m=p$. We define an $m \times m$ matrix

\begin{equation*}
\label{eq:matix_vector_relative_degree}
 A(x):=
  \left[
  \begin{array}{ccc}
      L_{g_1}L_{f}^{r_1 -1}h_1(x)& \cdots & L_{g_m}L_{f}^{r_1 -1}h_1(x)\\
      L_{g_1}L_{f}^{r_2 -1}h_2(x)& \cdots & L_{g_m}L_{f}^{r_2 -1}h_2(x)\\
      \vdots & \ddots & \vdots \\
      L_{g_1}L_{f}^{r_m -1}h_m(x)& \cdots & L_{g_m}L_{f}^{r_m -1}h_m(x)\\
  \end{array} \right].
\end{equation*}
The system has a vector relative degree of $\{r_1, \dots r_m \}$ at a point $x_0$ if
\begin{enumerate}
  \item $L_{g_j} L_f ^k h_i(x)=0,$ for all $1\leq j \leq m $ for all $ k<r_i-1$ for all $1\leq i \leq m$ and for all $x$ in a neighborhood of $x_0$, and
  \item  the matrix $A(x)$ is nonsingular at $x=x_0$.
\end{enumerate}
\end{definition}}

\begin{proposition}[Proposition 8.1, \cite{bhat2005geometric}]\label{prop:linearfinitetime}
Let $k_1,..., k_{n} > 0$ be such that the polynomial $s^n + k_ns^{n-1} + \cdots + k_{2}s + k_{1}$ is Hurwitz, and consider the system
\begin{equation}\label{eq:linearsystem}
    \begin{aligned}
    \dot{x}_{1} = x_{2},\\
    \vdots\\
    \dot{x}_{n - 1} = \dot{x}_{n},\\
    \dot{x}_{n} = u.
\end{aligned}
\end{equation}
There exists $\epsilon\in (0, 1)$ such that, for every $\alpha\in (1 - \epsilon, 1)$, the origin is a globally finite-time-stable equilibrium for the system in (\ref{eq:linearsystem}) under the feedback
$$
u = -k_1 \sign x_{1} |x_{1}|^{\alpha} - \cdots - k_{n}\sign x_{n} |x_{n}|^{\alpha_{n}},
$$
where $\alpha_{1},..., \alpha_{n} $ satisfy
\begin{equation}\label{eq:xicontrol}
    \alpha_{i - 1} = \frac{\alpha_{i}\alpha_{i + 1}}{2\alpha_{i + 1}- \alpha_{i}}, i = 2, ..., n,
\end{equation}
with $\alpha_{n + 1} = 1$ and $\alpha_{n} = \alpha$.
\end{proposition}
\begin{theorem}[Theorem 2 in \cite{XiaBelCas2022}]
    Given the reference profile $\eta_{2}^{\text{ref}}\in \reals$ and $\eta_{3}^{\text{ref}}\in \reals$ and the initial $\eta_{0}\in \reals^{6}$, if  If Problem 1 is initially feasible and the CBF constraint
in (24) corresponding to (22) does not conflict with both the control
bounds (2) and (18) at the same time, any controller $u\in$
guarantees the feasibility of problem (9), subject to (10)–(12).
\end{theorem}}